%% file: main.tex
\documentclass[10pt,twocolumn,letterpaper]{article}

\input{math_commands.tex}

\usepackage[pagenumbers]{cvpr} 

\usepackage{graphicx}
\usepackage{amsmath}
\usepackage{amssymb}
\usepackage{booktabs}

\usepackage{amsthm}

\newtheorem{corollary}{Corollary}
\newtheorem{proposition}{Proposition}
\newtheorem*{remark}{Remark}

\usepackage[pagebackref,breaklinks,colorlinks]{hyperref}

\usepackage[capitalize]{cleveref}
\crefname{section}{Sec.}{Secs.}
\Crefname{section}{Section}{Sections}
\Crefname{table}{Table}{Tables}
\crefname{equation}{}{}

\def\cvprPaperID{****} 
\def\confName{CVPR}
\def\confYear{2023}

\begin{document}

\title{OPERA: Omni-Supervised Representation Learning with Hierarchical Supervisions}

\newcommand*\samethanks[1][\value{footnote}]{\footnotemark[#1]}

\author{Chengkun Wang$^{1,2,}$\footnotetext{sadfsadfa}\thanks{Equal contribution.}\quad Wenzhao Zheng$^{1,2,}$\samethanks\quad Zheng Zhu$^3$\quad Jie Zhou$^{1,2}$\quad Jiwen Lu$^{1,2,}$\thanks{Corresponding author.} \\
$^1$Beijing National Research Center for Information Science and Technology, China \\
$^2$Department of Automation, Tsinghua University, China \quad\quad $^3$PhiGent Robotics \\
\texttt{\{wck20,zhengwz18\}@mails.tsinghua.edu.cn; zhengzhu@ieee.org;} \\
\texttt{\{jzhou,lujiwen\}@tsinghua.edu.cn}
}
\maketitle

\input{chapters/0_abstract.tex}
\input{chapters/1_introduction.tex}

\input{chapters/2_related_work.tex}

\input{chapters/3_proposed_approach.tex}
\input{chapters/4_experiment.tex}
\input{chapters/5_conclusion.tex}

\appendix
\input{chapters/appendix.tex}

{\small

\input{reference.bbl}
\bibliographystyle{ieee_fullname}
}

\end{document}

%% file: math_commands.tex
\usepackage{amsmath,amsfonts,bm}

\def\1{\bm{1}}

\def\vy{{\bm{y}}}

\def\mI{{\bm{I}}}

\def\mW{{\bm{W}}}

\DeclareMathAlphabet{\mathsfit}{\encodingdefault}{\sfdefault}{m}{sl}
\SetMathAlphabet{\mathsfit}{bold}{\encodingdefault}{\sfdefault}{bx}{n}



%% file: chapters/0_abstract.tex
\begin{abstract}
The pretrain-finetune paradigm in modern computer vision facilitates the success of self-supervised learning, which achieves better transferability than supervised learning.
However, with the availability of massive labeled data, a natural question emerges: \emph{how to train a better model with \textbf{both} self and full supervision signals?}
In this paper, we propose \textbf{O}mni-su\textbf{PE}rvised \textbf{R}epresentation  le\textbf{A}rning with hierarchical supervisions (\textbf{OPERA}) as a solution.
We provide a unified perspective of supervisions from labeled and unlabeled data and propose a unified framework of fully supervised and self-supervised learning.
We extract a set of hierarchical proxy representations for each image and impose self and full supervisions on the corresponding proxy representations. 
Extensive experiments on both convolutional neural networks and vision transformers demonstrate the superiority of OPERA in image classification, segmentation, and object detection.
Code is available at: \url{https://github.com/wangck20/OPERA}.

\end{abstract}

%% file: chapters/1_introduction.tex
\section{Introduction} 
Learning good representations is a significant yet challenging task in deep learning~\cite{chen2021exploring,zheng2021deepr,he2020momentum}.
Researchers have developed various ways to adapt to different supervisions, such as fully supervised~\cite{oh2018modeling,kim2020proxy,wang2016joint,verma2019manifold}, self-supervised~\cite{wang2015unsupervised,ye2020probabilistic,grill2020bootstrap,chen2020simple}, and semi-supervised learning~\cite{xu2021end,zhang2021flexmatch,wang2022np}.
They serve as fundamental procedures in various tasks including image classification~\cite{deng2019arcface,zhang2018mixup,yun2019cutmix}, semantic segmentation~\cite{grill2020bootstrap,strudel2021segmenter}, and object detection~\cite{he2017mask,yang2019reppoints,carion2020end}.

\begin{figure}[t]
\centering
\includegraphics[width=0.475\textwidth]{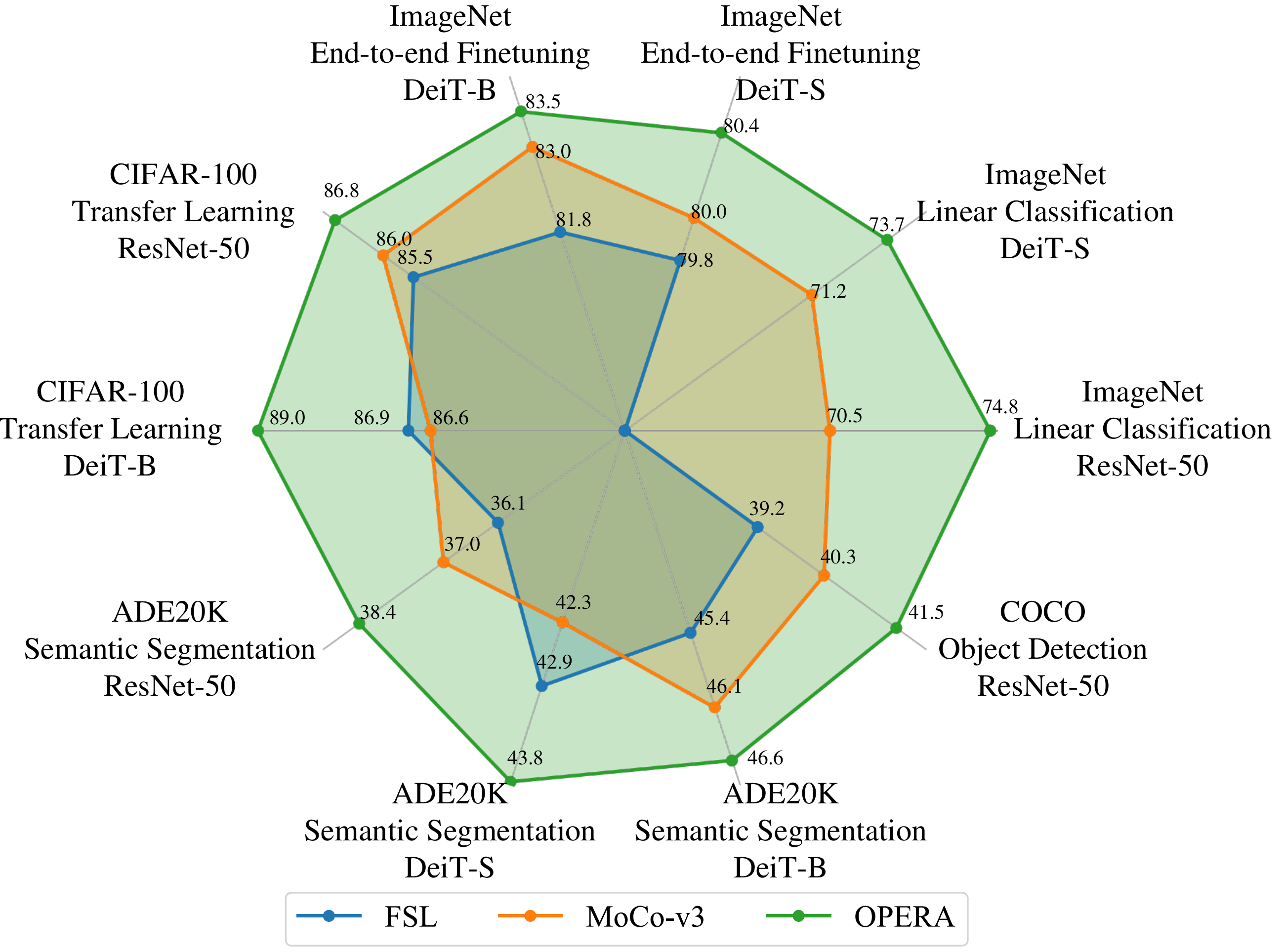}
\vspace{-7mm}
\caption{The proposed OPERA outperforms both fully supervised and self-supervised counterparts on various downstream tasks.
}
\label{fig:radar}
\vspace{-7mm}
\end{figure}

Fully supervised learning (FSL) has always been the default choice for representation learning, which learns from discriminating samples with different ground-truth labels. 
However, this dominance begins to fade with the rise of the pretrain-finetune paradigm in modern computer vision.
Under such a paradigm, researchers usually pretrain a network on a large dataset first and then transfer it to downstream tasks~\cite{he2021masked,chu2021twins,he2020momentum,chen2021exploring}.
This advocates transferability more than discriminativeness of the learned representations.
This preference nurtures the recent success of self-supervised learning (SSL) methods with contrastive objective~\cite{he2020momentum,xie2021detco,grill2020bootstrap,chen2020simple,wang2022contrastive}.
They require two views (augmentations) of the same image to be consistent and distinct from other images in the representation space.
This instance-level supervision is said to obtain more general and thus transferable representations~\cite{ericsson2021well,islam2021broad}.
The ability to learn without human-annotated labels also greatly popularizes self-supervised contrastive learning.
Despite its advantages, we want to explore \emph{whether combining self-supervised signals\footnote{We mainly focus on self-supervised contrastive learning. In the rest of the paper, we use self-supervised learning to refer to self-supervised contrastive learning unless otherwise specified for simplicity.} with fully supervised signals further improves the transferability}, given the already availability of massive annotated labels~\cite{russakovsky2015imagenet,lin2014microsoft,abu2016youtube,caesar2020nuscenes}.

\begin{figure*}[t]
\centering
\includegraphics[width=0.9\textwidth]{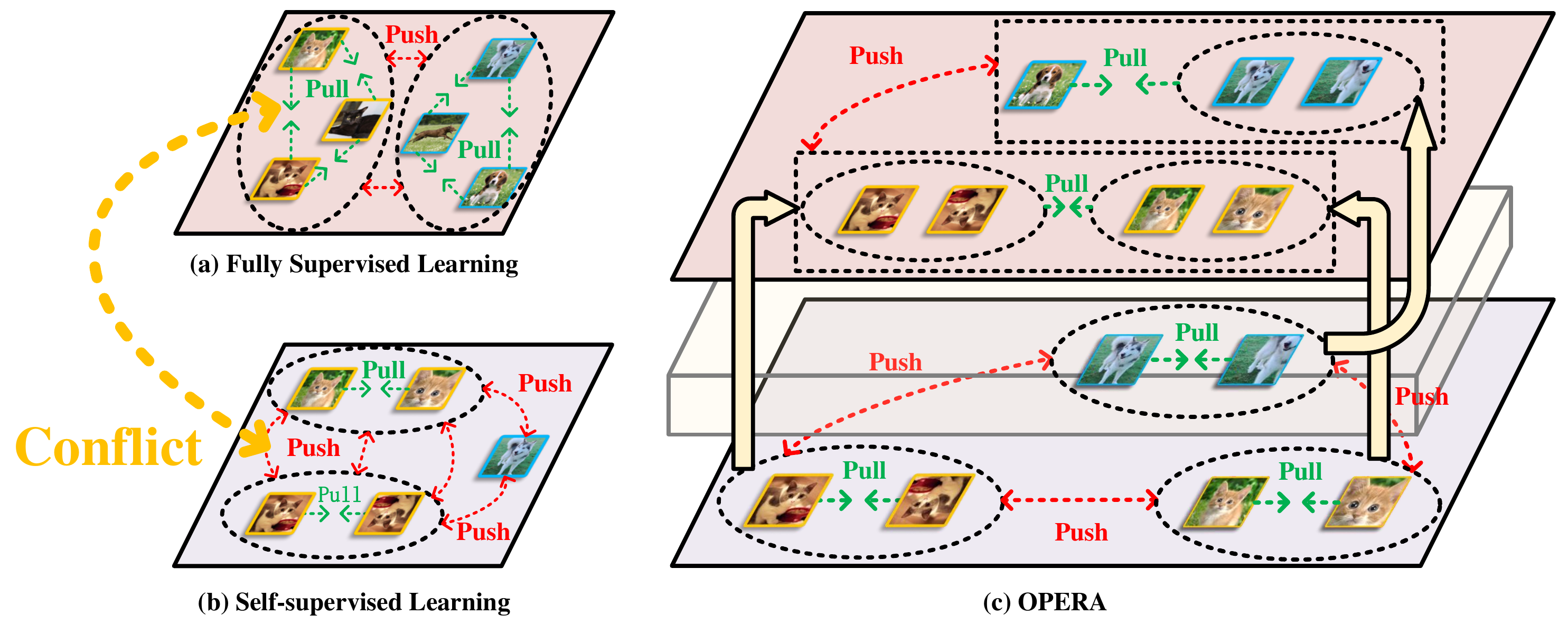}
\vspace{-4mm}
\caption{
Comparisons of different learning strategies. 
Fully supervised learning (a) and self-supervised learning (b) constrain images at the class level and instance level, respectively. 
They conflict with each other for different images from the same class.
OPERA imposes hierarchical supervisions on hierarchical spaces and uses a transformation to resolve the supervision conflicts.
} 
\vspace{-6mm}
\label{fig:comparison}
\end{figure*}

We find that a simple combination of the self and full supervisions results in contradictory training signals.
To address this, in this paper, we provide \textbf{O}mni-su\textbf{PE}rvised \textbf{R}epresentation  le\textbf{A}rning with hierarchical supervisions (\textbf{OPERA}) as a solution, as demonstrated in \Cref{fig:comparison}. 
We unify full and self supervisions in a similarity learning framework where they differ only by the definition of positive and negative pairs.
Instead of directly imposing supervisions on the representations, we extract a hierarchy of proxy representations to receive the corresponding supervision signals. 
Extensive experiments are conducted with both convolutional neural networks~\cite{he2016deep} and vision transformers~\cite{dosovitskiy2020image} as the backbone model.
We pretrain the models using OPERA on ImageNet-1K~\cite{russakovsky2015imagenet} and then transfer them to various downstream tasks to evaluate the transferability.
We report image classification accuracy with both linear probe and end-to-end finetuning on ImageNet-1K.
We also conduct experiments when transferring the pretrained model to other classification tasks, semantic segmentation, and object detection. 
Experimental results demonstrate consistent improvements over FSL and SSL on all the downstream tasks, as shown in \Cref{fig:radar}.
Additionally, we show that OPERA outperforms the counterpart methods even with fewer pretraining epochs (e.g., fewer than 150 epochs), demonstrating good data efficiency.

%% file: chapters/2_related_work.tex
\section{Related Work}

\textbf{Fully Supervised Representation Learning.}
Fully supervised representation learning (FSL) utilizes the ground-truth labels of data to learn a discriminative representation space.
The general objective is to maximize the discrepancies of representations from different categories and minimize those from the same class.
The softmax loss is widely used for FSL~\cite{he2016deep,liu2021swin, deng2019arcface, wang2018cosface}, and various loss functions are further developed in deep metric learning~\cite{kim2020proxy,wang2019multi,hu2014discriminative,movshovitz2017no,teh2020proxynca++}.

As fully supervised objectives entail strong constraints, the learned representations are usually more suitable for the specialized classification task and thus lag behind on transferability~\cite{zhao2020makes,ericsson2021well,islam2021broad}.
To alleviate this, many works devise various data augmentation methods to expand the training distribution~\cite{zhang2018mixup,kim2020puzzle,chen2022transmix,venkataramanan2022alignmixup}.
Recent works also explore adding more layers after the representation to avoid direct supervision~\cite{vo2019generalization, wang2022revisiting}.
Differently, we focus on effectively combining self and full supervisions to improve transferability.

\textbf{Self-supervised Representation Learning.}
Self-supervised representation learning (SSL) attracts increasing attention in recent years due to its ability to learn meaningful representation without human-annotated labels.
The main idea is to train the model to perform a carefully designed label-free pretext task.
Early self-supervised learning methods devised various pretext tasks including image restoration~\cite{vincent2008extracting,zhang2016colorful,pathak2016context}, prediction of image rotation~\cite{gidaris2018unsupervised}, and solving jigsaw
puzzles~\cite{noroozi2016unsupervised}.
They achieve fair performance but still cannot equal fully supervised learning until the arise of self-supervised contrastive learning~\cite{he2020momentum,chen2020simple,grill2020bootstrap}.
The pretext task of contrastive learning is instance discrimination, i.e., to identify different views (augmentations) of the same image from those of other images.
Contrastive learning methods~\cite{chen2021exploring,xie2021detco,wang2022cp2,xie2020pointcontrast,liu2020self,chen2021multisiam,hou2021pri3d,liang2021exploring} demonstrate even better transferability than fully supervised learning, resulting from their focus on lower-level and thus more general features~\cite{zhao2020makes,ericsson2021well,islam2021broad}.
Very recently, masked image modeling (MIM)~\cite{he2021masked,zhou2021ibot,xie2022simmim} emerges as a strong competitor to contrastive learning, which trains the model to correctly predict the masked parts of the input image.
In this paper, we mainly focus on contrastive learning in self-supervised learning.
Our framework can be extended to other pretext tasks by inserting a new task space in the hierarchy.

\textbf{Omni-supervised Representation Learning:}
It is worth mentioning that some existing studies have attempted to combine FSL and SSL~\cite{radosavovic2018data,nayman2022diverse,wei2022can}.
Radosavovic~et el.~\cite{radosavovic2018data} first trained an FSL model and then performed knowledge distillation on unlabeled data. 
Wei~et el.~\cite{wei2022can} adopted an SSL pretrained model to generate instance labels and compute an overall similarity to train a new model.
Nayman~et el.~\cite{nayman2022diverse} proposed to finetune an SSL pretrained model using ground-truth labels in a controlled manner to enhance its transferability.
Nevertheless, they do not consider the hierarchical relations between the self and full supervision. 
Also, they perform SSL and FSL sequentially in separate stages.
Differently, OPERA unifies them in a universal perspective and imposes the supervisions on different levels of the representations. 
Our framework can be trained in an end-to-end manner efficiently with fewer epochs.

%% file: chapters/3_proposed_approach.tex
\section{Proposed Approach}

In this section, we first present a unified perspective of self-supervised learning (SSL) and fully supervised learning (FSL) under a similarity learning framework. 
We then propose OPERA to impose hierarchical supervisions on the corresponding hierarchical representations for better transferability. 
Lastly, we elaborate on the instantiation of the proposed OPERA framework.

\subsection{Unified Framework of Similarity Learning}
 Given an image space $\mathcal{X} \subset  \mathcal{R}^{H\times W \times C}$, deep representation learning trains a deep neural network as the map to their representation space $\mathcal{Y} \subset \mathcal{R}^{D \times 1}$.
 Fully supervised learning and self-supervised learning are two mainstream representation learning approaches in modern deep learning. 
FSL utilizes the human-annotated labels as explicit supervision to train a discriminative classifier. 
Differently, SSL trains models without ground-truth labels.
The widely used contrastive learning (\emph{e.g.}, MoCo-v3~\cite{chen2021empirical}) obtains meaningful representations by maximizing the similarity between random augmentations of the same image.

Generally, FSL and SSL differ in both the supervision form and optimization objective.
To integrate them, we first provide a unified similarity learning framework to include both  training objectives:
\begin{equation}\label{equ:similarity learning}
\begin{aligned}
J(\mathcal{Y}, \mathcal{P}, \mathcal{L}) &= \sum_{\mathbf{y} \in \mathcal{Y}, \mathbf{p} \in \mathcal{P}, l \in \mathcal{L}} [ -w_p \cdot I(l_{\mathbf{y}}, l_{\mathbf{p}}) \cdot s(\mathbf{y},\mathbf{p}) \\
&+ w_n \cdot (1-I(l_{\mathbf{y}}, l_{\mathbf{p}})) \cdot s(\mathbf{y},\mathbf{p})],
\end{aligned}
\end{equation}
where  $w_p \geq 0$ and $w_n \geq 0$ denote the coefficients of positive and negative pairs, $l_{\mathbf{y}}$ and $l_{\mathbf{p}}$ are the labels of the samples, and $s(\mathbf{y},\mathbf{p})$ defines the pairwise similarity between $\mathbf{y}$ and $\mathbf{p}$.
$I(a,b)$ is an indicator function which outputs 1 if $a=b$ and 0 otherwise.
$\mathcal{L}$ is the label space, and $\mathcal{P}$ can be the same as $\mathcal{Y}$, a transformation of $\mathcal{Y}$, or a learnable class prototype space.
For example, to obtain the softmax objective widely employed in FSL~\cite{he2016deep,touvron2021training}, we can set: 
\begin{equation}\label{equ:softmax}
	w_p=1, w_n=\frac{exp(s(\mathbf{y},\mathbf{p}))}{\sum_{l_{\mathbf{p'}} \neq l_{\mathbf{y}}} exp(s(\mathbf{y},\mathbf{p'}))},
\end{equation}
where $s(\mathbf{y},\mathbf{p}) = \mathbf{y}^T \cdot \mathbf{p}$, and $\mathbf{p}$ is the row vector in the classifier matrix $\mathbf{W}$.
For the InfoNCE loss used in contrastive learning~\cite{van2018representation,he2020momentum,khosla2020supervised}, we set:
\begin{equation}\label{equ:infonce}
\begin{aligned}
	w_p=\frac{1}{\tau}\frac{\sum_{l_{l_{\mathbf{p'}} \neq \mathbf{y}}} exp(s(\mathbf{y},\mathbf{p'})/\tau)}{exp(s(\mathbf{y},\mathbf{p})/\tau) + \sum_{{l_{\mathbf{p'}} \neq l_{\mathbf{y}}}} exp(s(\mathbf{y},\mathbf{p'})/\tau)}, \\
	w_n=\frac{1}{\tau}\frac{ exp(s(\mathbf{y},\mathbf{p})/\tau)}{exp(s(\mathbf{y},\mathbf{p})/\tau) + \sum_{l_{\mathbf{p'}} \neq l_{\mathbf{y}}} exp(s(\mathbf{y},\mathbf{p'})/\tau)}
	\end{aligned}
\end{equation}
where $\tau$ is the temperature hyper-parameter. 
See \cref{appendix:proof of softmax and infonce} for more details.

Under the unified training objective \eqref{equ:similarity learning}, the main difference between FSL and SSL lies in the definition of the label space $\mathcal{L}^{full}$ and $\mathcal{L}^{self}$.
For the labels $l^{full} \in \mathcal{L}^{full}$ in FSL, $l^{full}_i=l^{full}_j$ only if they are from the same ground-truth category.
For the labels $l^{self} \in \mathcal{L}^{self}$ in SSL, $l^{self}_i=l^{self}_j$ only if they are the augmented views of the same image.

\begin{figure*}[t]
\centering
\includegraphics[width=0.99\textwidth]{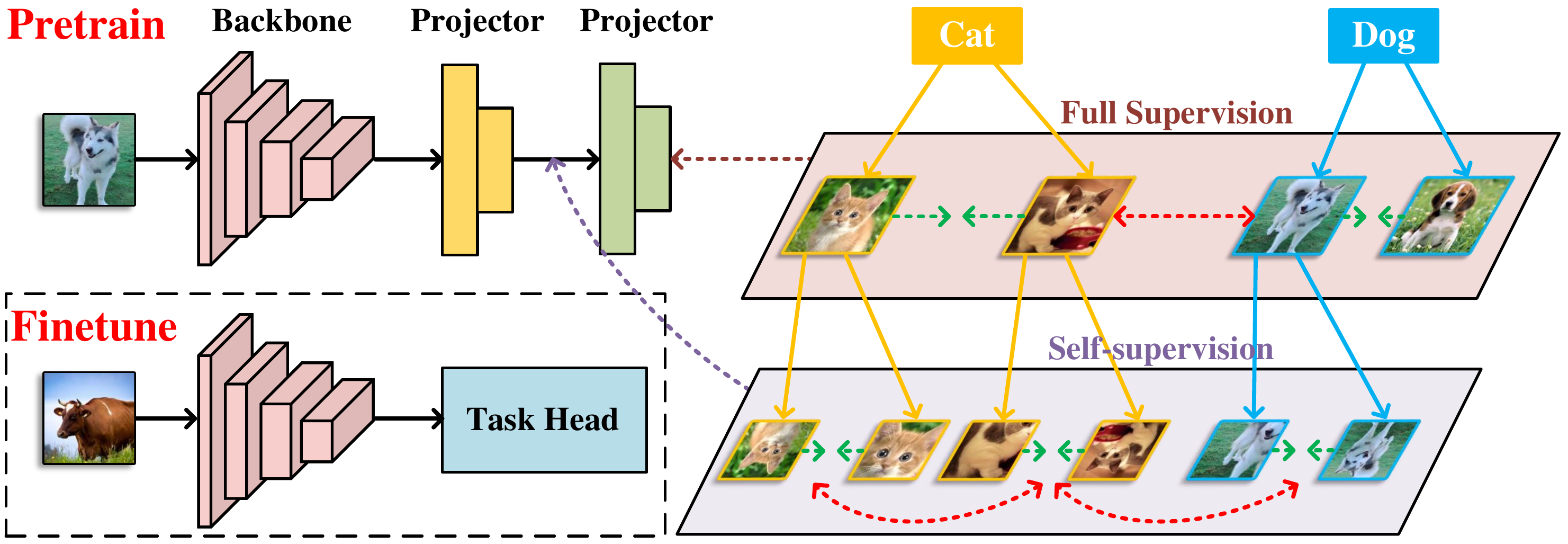}
\vspace{-4mm}
\caption{
An illustration of the proposed OPERA framework. 
We impose perform SSL and FSL on the corresponding proxy representations, respectively.
OPERA combines both supervisions to balance instance-level and class-level information for the backbone in an end-to-end manner.
} 
\vspace{-7mm}
\label{fig:framework}
\end{figure*}

\subsection{Hierarchical Supervisions on Hierarchical Representations}
With the same formulation of the training objective, a naive way to combine the two training signals is to simply add them:
\begin{small}
\begin{equation}\label{equ:naive}
\begin{aligned}
J^{naive}(\mathcal{Y}, \mathcal{P}, \mathcal{L}) &=  \sum_{\mathbf{y} \in \mathcal{Y}, \mathbf{p} \in \mathcal{P}, l \in \mathcal{L}}  [ -w_p^{self} \cdot I(l_{\mathbf{y}}^{self}, l_{\mathbf{p}}^{self}) \cdot s(\mathbf{y},\mathbf{p}) \\
&+ w_n^{self} \cdot (1-I(l_{\mathbf{y}}^{self}, l_{\mathbf{p}}^{self})) \cdot s(\mathbf{y},\mathbf{p}) \\
 &-w_p^{full} \cdot I(l_{\mathbf{y}}^{full}, l_{\mathbf{p}}^{full}) \cdot s(\mathbf{y},\mathbf{p})\\
 &+ w_n^{full} \cdot (1-I(l_{\mathbf{y}}^{full}, l_{\mathbf{p}}^{full})) \cdot s(\mathbf{y},\mathbf{p})]. 
\end{aligned}
\end{equation}
\end{small}

For $\mathbf{y}$ and $\mathbf{p}$ from the same class, i.e., $I(l_{\mathbf{y}}^{self}, l_{\mathbf{p}}^{self}) = 0$ and $I(l_{\mathbf{y}}^{full}, l_{\mathbf{p}}^{full}) = 1$, the training loss is:
\begin{equation}\label{equ:naive}
J^{naive}(\mathbf{y}, \mathbf{p}, \mathbf{l}) =  (w_n^{self} - w_p^{full}) \cdot s(\mathbf{y},\mathbf{p}). 
\end{equation}
This indicates the two training signals are contradictory and may neutralize each other.
This is particularly harmful if we adopt similar loss functions for fully supervised and self-supervised learning, i.e., $w_n^{self} \approx w_p^{full}$, and thus $J^{naive}(\mathbf{y}, \mathbf{p}, \mathbf{l}) \approx 0$.

Existing methods~\cite{nayman2022diverse,wei2022can,wang2022revisiting} address this by subsequently imposing the two training signals.
They tend to first obtain a self-supervised pretrained model and then use the full supervision to tune it.
Differently, we propose a more efficient way to adaptively balance the two weights so that we can simultaneously employ them:
\begin{equation}\label{equ:adaptive}
J^{adap}(\mathbf{y}, \mathbf{p}, \mathbf{l}) =  (w_n^{self} \cdot \alpha - w_p^{full} \cdot \beta) \cdot s(\mathbf{y},\mathbf{p}),
\end{equation}
where $\alpha$ and $\beta$  are modulation factors that can be dependent on $\mathbf{y}$ and $\mathbf{p}$ for more flexibility.
However, it remains challenging to design the specific formulation of $\alpha$ and $\beta$.

Considering that the two label spaces are entangled and demonstrate a hierarchical structure:
\begin{equation}
I(l_{\mathbf{y}}^{self}, l_{\mathbf{p}}^{self}) = 1 \implies  I(l_{\mathbf{y}}^{full}, l_{\mathbf{p}}^{full}) = 1, 
\end{equation}
i.e., the two augmented views of the same image must share the same category label, 
we transform the image representation into proxy representations in an instance space and a class space to construct a hierarchical structure.
Formally, we apply two transformations $\mathcal{Y}$ sequentially: 
\begin{equation}
\mathcal{Y}^{self}=g(\mathcal{Y}), \quad \mathcal{Y}^{full}=h(\mathcal{Y}^{self}),
\end{equation}
where $g(\cdot)$ and $h(\cdot)$ denote the mapping functions.
We extract the class representations following the instance representations since full supervision encodes higher-level features than self-supervision.

We then impose the self and full supervision on the instance space and class space, respectively, to formulate the overall training objective for the proposed OPERA:
\begin{equation}\label{equ:hierarchical}
\begin{aligned}
J^{O}(\mathcal{Y},\mathcal{P},\mathcal{L})&=J^{self}(\mathcal{Y}^{self},\mathcal{P}^{self},\mathcal{L}^{self}) \\
&+ J^{full}(\mathcal{Y}^{full},\mathcal{P}^{full},\mathcal{L}^{full}).  
\end{aligned}
\end{equation}
We will show in the next subsection that this objective naturally implies \eqref{equ:adaptive}, which implicitly and adaptively balances self and full supervisions in the representation space.

\subsection{Omni-supervised Representation Learning}

To effectively combine the self and full supervision to learn representations, OPERA further extracts a set of proxy representations hierarchically to receive the corresponding training signal, as illustrated in \Cref{fig:framework}.
Despite its simplicity and efficiency, it is not clear how it achieves balances between the two supervision signals and how it resolves the contradiction demonstrated in \eqref{equ:naive}.

To thoroughly understand the effect of \eqref{equ:hierarchical} on the image representations, we project it back on the representation space $\mathcal{Y}$ and obtain an equivalent training objective in $\mathcal{Y}$.

\begin{proposition} \label{prop1}
Assume using linear projection as the transformation between representation spaces. $g(\vy) = \mW_g \vy$ and $h(\vy) = \mW_h \vy$, where $\mW_g$ and $\mW_h$ are learnable parameters.
Optimizing \eqref{equ:hierarchical} is equivalent to optimizing the following objective on the original representation space $\mathcal{Y}$:
\begin{equation}\label{equ:equivalent}
\begin{aligned}
J(\mathcal{Y},\mathcal{P},\mathcal{L}) &= \sum_{\mathbf{y} \in \mathcal{Y}, \mathbf{p} \in \mathcal{P}, l \in \mathcal{L}} [I(l_{\mathbf{y}}^{self},l_{\mathbf{p}}^{self})\cdot I(l_{\mathbf{y}}^{full},l_{\mathbf{p}}^{full})\\
&\cdot (-w_p^{self}\alpha(\mW_g)-w_p^{full}\beta(\mW_g,\mW_h))\cdot s(\mathbf{y},\mathbf{p}) \\
&+ (1-I(l_{\mathbf{y}}^{self},l_{\mathbf{p}}^{self}))\cdot I(l_{\mathbf{y}}^{full},l_{\mathbf{p}}^{full})\\
&\cdot (w_n^{self}\alpha(\mW_g)-w_p^{full}\beta(\mW_g,\mW_h))\cdot s(\mathbf{y},\mathbf{p}) \\
&+ (1-I(l_{\mathbf{y}}^{self},l_{\mathbf{p}}^{self}))\cdot (1-I(l_{\mathbf{y}}^{full},l_{\mathbf{p}}^{full}))\\
&\cdot(w_n^{self}\alpha(\mW_g) + w_n^{full}\beta(\mW_g,\mW_h))\cdot s(\mathbf{y},\mathbf{p})],
\end{aligned}
\end{equation}
where $\alpha(\mW_g)$ and $\beta(\mW_g,\mW_h)$ are scalars related to the transformation parameters.
\end{proposition}
We give detailed proof in~\cref{appendix:proof of prop}.
\begin{remark}
    \cref{prop1} only considers the case without activation functions.
    We conjecture that the mappings $g(\cdot)$ and $h(\cdot)$ only influence the form of $\beta(\cdot, \cdot)$ without altering the final conclusion. 
\end{remark}

\cref{prop1} induces two corollaries as proved in~\cref{appendix:proof of coro1} and~\cref{appendix:proof of coro2}.

\begin{corollary} \label{coro1}
The loss weight $w$ on a pair of samples $(\mathbf{y},\mathbf{p})$ satisfies:
\begin{small}
\begin{equation}
\begin{aligned}
	w(l_{\mathbf{y}}^{self}=l_{\mathbf{p}}^{self},l_{\mathbf{y}}^{full}=l_{\mathbf{p}}^{full}) &\leq w(l_{\mathbf{y}}^{self}\neq l_{\mathbf{p}}^{self},l_{\mathbf{y}}^{full}=l_{\mathbf{p}}^{full}) \\
	&\leq w (l_{\mathbf{y}}^{self}\neq l_{\mathbf{p}}^{self}, l_{\mathbf{y}}^{full}\neq l_{\mathbf{p}}^{full}).
\end{aligned}
\end{equation}
\end{small}
\end{corollary}

\begin{corollary} \label{coro2}
We resolve the contradictory in~\eqref{equ:naive} by adaptively adjusting the loss weight by 
\begin{equation}
w_n^{self} \cdot  \alpha(\mW_g) - w_p^{full} \cdot  \beta(\mW_g,\mW_h). 
\end{equation}
\end{corollary}

\cref{coro1} ensures that the learned representations are consistent with how humans perceive the similarities of images, i.e., the similarities between different images of the same class should be larger than those between images of different classes but smaller than those between the views of the same images.
\cref{coro2} demonstrates the ability of OPERA to adaptively balance the training signals of self and full supervisions.

OPERA can be trained in an end-to-end manner using both self and full supervisions.
We extract proxy representations in hierarchical spaces to receive the corresponding training signals.
For inference, we discard the proxy representations and directly add the task head on the image representation space $\mathcal{Y}$.
\subsection{Instantiation of OPERA}
We present the instantiation of the proposed omni-supervised representation learning with hierarchical supervisions. In the pretraining procedure, we extract hierarchical proxy representations for each image $\mathbf{x}_i$ in our model, denoted as $\{\mathbf{y}_i^{self}, \mathbf{y}_i^{full}\}$. We conduct self-supervised learning with the instance-level label $l_i^{self}$ on the instance-level representation $\mathbf{y}_i^{self}$ and the class-level label $l_i^{full}$ is imposed on $\mathbf{y}_i^{full}$. The overall objective of our framework follows~\eqref{equ:hierarchical} and OPERA can be optimized in an end-to-end manner. During finetuning, the downstream task head is directly applied to the learned representations $\mathcal{Y}$. The transfer learning includes image classification and other dense prediction tasks such as semantic segmentation.

In this paper, we apply OPERA to MoCo-v3~\cite{chen2021empirical} by instantiating $\mathcal{Y}^{self}$ as the output of the online predictor and the target predictor denoted as $\mathcal{Y}_q^{self}$ and $\mathcal{Y}_k^{self}$, respectively. Additionally, $J(\mathcal{Y}^{self},\mathcal{L}^{self})$ is the widely-used InfoNCE loss~\cite{van2018representation}. Furthermore, we employ an extra MLP block that explicitly connects to the online predictor to obtain $\mathcal{Y}^{full}$ and fix the output dimension to the class number of the pretrained dataset (\emph{e.g.}, 1,000 for ImageNet). We then introduce full supervision on $\mathcal{Y}^{full}$ with the Softmax loss. The overall objective based on MoCo-v3 is as follows:
\begin{equation}
\begin{aligned}
&J_m(\mathcal{Y},\mathcal{L})=\frac{1}{N}\sum_{i=1}^{N}[-log\frac{ exp(\mathbf{y}_{i,l_i}^{full})}{\sum_{j\neq l_{i}}exp(\mathbf{y}_{i,j}^{full})}\\
&-log\frac{exp(\mathbf{y}_{q,i}^{self}\cdot \mathbf{y}_{k,i}^{self} /\tau)}{exp(\mathbf{y}_{q,i}\cdot \mathbf{y}_{k,i} /\tau) + \sum_{j\neq i}exp(\mathbf{y}_{q,i}\cdot \mathbf{y}_{k,j} /\tau)}]
\end{aligned}
\end{equation}
where $\mathbf{y}_{i,j}^{full}$ denotes the $j$th component of $\mathbf{y}_{i}^{full}$. In addition, we also adopt the stop-gradient operation and the momentum update to the target network following~\cite{he2020momentum}. Therefore, the proposed OPERA framework preserves the instance-level information in MoCo-v3 to prevent damaging the transferability of the model. Furthermore, OPERA involves class-level knowledge with the class-level full supervision, which further boosts the performance of the learned representations.

%% file: chapters/4_experiment.tex
\newcommand{\tablesize}{\small}

\section{Experiments}
In this section, we conducted extensive experiments to evaluate the performance of our OPERA framework. 
We pretrained the network using OPERA on the ImageNet-1K~\cite{russakovsky2015imagenet} (IN) dataset and then evaluated its performance on different tasks.
We provide in-depth ablation studies to analyze the effectiveness of OPERA.
All experiments were conducted with PyTorch~\cite{paszke2019pytorch} using RTX 3090 GPUs.

\subsection{Experimental Setup}
\textbf{Datasets.}
We pretrain our model on the training set of ImageNet-1K~\cite{russakovsky2015imagenet} containing 1,200,000 samples of 1,000 categories.
We evaluate the linear probe and end-to-end finetuning performance on the validation set consisting of 50,000 images.
For transferring to other classification tasks, we use CIFAR-10~\cite{krizhevsky2009learning}, CIFAR-100~\cite{krizhevsky2009learning}, Oxford Flowers-102~\cite{nilsback2008automated}, and Oxford-IIIT-Pets~\cite{parkhi2012cats}. 
For other downstream tasks, we use ADE20K~\cite{zhou2019semantic} for semantic segmentation and COCO~\cite{lin2014microsoft} for object detection and instance segmentation.

\textbf{Implementation Details.}We mainly applied our OPERA to MoCo-v3~\cite{chen2021empirical}.
We added an extra MLP block after the predictor of the online network, which is composed of two fully-connected layers with a batch normalization layer and a ReLU layer. 
The hidden dimension of the MLP block was set to $256$ while the output dimension was $1,000$. 
We trained ResNet50~\cite{he2016deep} (R50) and ViTs~\cite{touvron2021training,dosovitskiy2020image} (ViT-S and ViT-B) as our backbone with a batch size of $1024$, $2048$, and $4096$.
We adopted LARS~\cite{you2017large} as the optimizer for R50 and AdamW~\cite{loshchilov2018decoupled} for ViT. 
We set the other settings the same as the original MoCo-v3 for fair comparisons.
In the following experiments, $\dag$ denotes our reproduced results with the same settings and BS denotes the batch size.
P.T and F.T denote the pretraining and finetuning epochs, respectively.
The bold number highlights the improvement of OPERA compared with the associated method, and the red number indicates the best performance.

\begin{table}[t] \tablesize 
    \centering
    \caption{Top-1 and top-5 accuracies (\%) under the linear classification protocol on ImageNet. }
    \vspace{-3mm}
    \setlength\tabcolsep{2pt}
    \begin{tabular}{lcccccc}
    \toprule
    Method & BS & P.T. & F.T. & Backbone & Top-1 Acc & Top-5 Acc \\
    \midrule
    MoCo-v1 & 256 & 200 & 100 & R50 & 60.6 & - \\
    MoCo-v2 & 256 & 200 & 100 & R50 & 67.5 & - \\
    MoCo-v2 & 256 & 800 & 100 & R50 & 71.1 & - \\
    SimCLR & 4096 & 100 & 1000 & R50 & 69.3 & 89.0 \\
    SimSiam & 256 & 800 & 100 & R50 & 71.3 & - \\
    BYOL & 4096 & 1000 & 80 & R50 & 74.3 & 91.6 \\
    \midrule
    MoCo-v3$\dag$ & 1024 & 300 & 90 & R50 & 70.5 & 90.0 \\
    OPERA & 1024 & 150 & 90 & R50 & \textbf{73.7} & \textbf{91.2} \\
    OPERA & 1024 & 300 & 90 & R50 & \color{red}\textbf{74.8} & \color{red}\textbf{91.9} \\
    \midrule
    MoCo-v3$\dag$ & 1024 & 300 & 90 & ViT-S & 71.2 & 90.3 \\
    OPERA & 1024 & 150 & 90 & ViT-S & \textbf{72.7} & \textbf{90.7}\\
    OPERA & 1024 & 300 & 90 & ViT-S & \color{red}\textbf{73.7} & \color{red}\textbf{91.3}\\
    \bottomrule
    \end{tabular}
    \vspace{-4mm}
    \label{tab:linear classification}
\end{table}

\begin{table}[t] \tablesize
    \centering
    \caption{Top-1 and top-5 accuracies (\%) under the end-to-end finetuning protocol on ImageNet. }
    \vspace{-3mm}
    \setlength\tabcolsep{2.4pt}
    \begin{tabular}{lcccccc}
    \toprule
    Method & BS & P.T. & F.T. & Backbone & Top-1 Acc & Top-5 Acc \\
    \midrule
    Supervised & 1024 & - & 300 & ViT-S & 79.8 & 95.0 \\ 
    Supervised & 1024 & - & 300 & ViT-B & 81.8 & 95.6 \\
    DINO$\dag$ & 1024 & 300 & 300 & ViT-B & 82.8 & 96.3 \\
    \midrule
    MoCo-v3$\dag$ & 1024 & 300 & 100 & ViT-S & 78.8 & 94.6 \\
    OPERA & 1024 & 150 & 100 & ViT-S & \textbf{79.1} & \textbf{94.7}\\
    OPERA & 1024 & 300 & 100 & ViT-S & \textbf{80.0} & \textbf{95.1}\\
    \midrule
    MoCo-v3$\dag$ & 1024 & 300 & 150 & ViT-S & 79.1 & 94.6 \\
    OPERA & 1024 & 150 & 150 & ViT-S & \textbf{79.9} & \textbf{95.1}\\
    OPERA & 1024 & 300 & 150 & ViT-S & \textbf{80.4} & \textbf{95.3}\\
    \midrule
    MoCo-v3$\dag$ & 1024 & 300 & 200 & ViT-S & 80.0 & 95.2 \\
    OPERA & 1024 & 300 & 200 & ViT-S & \color{red}\textbf{80.8} & \color{red}\textbf{95.5}\\
    \midrule
    MoCo-v3$\dag$ & 1024 & 300 & 150 & ViT-B & 82.1 & 95.9 \\
    OPERA & 1024 & 150 & 150 & ViT-B & \textbf{82.4} & \textbf{96.0}\\
    OPERA & 1024 & 300 & 150 & ViT-B & \textbf{82.6} & \textbf{96.2}\\
    \midrule
    MoCo-v3$\dag$ & 2048 & 300 & 150 & ViT-B & 82.7 & 96.3 \\
    OPERA & 2048 & 150 & 150 & ViT-B & \textbf{82.8} & \textbf{96.3}\\
    OPERA & 2048 & 300 & 150 & ViT-B & \textbf{83.1} & \textbf{96.4}\\
    \midrule
    MoCo-v3$\dag$ & 4096 & 300 & 150 & ViT-B & 83.0 & 96.3 \\
    OPERA & 4096 & 150 & 150 & ViT-B & \textbf{83.2} & \textbf{96.4}\\
    OPERA & 4096 & 300 & 150 & ViT-B & \color{red}\textbf{83.5} & \color{red}\textbf{96.5}\\
    \bottomrule
    \end{tabular}
    \vspace{-7mm}
    \label{tab:Finetuning}
\end{table}

\subsection{Main Results}
\textbf{Linear Probe Evaluation on ImageNet.}
We evaluated OPERA using the linear probe protocol, where we trained a classifier on top of the frozen representation. 
We used the SGD~\cite{1985A} optimizer and fixed the batch size to 1024.
We set the learning rate to 0.1 for R50~\cite{he2016deep} and 3.0 for ViT-S~\cite{touvron2021training}.
The weight decay was 0 and the momentum of the optimizer was 0.9 for both architectures. 
We also compared OPERA with existing SSL methods including MoCo-v1~\cite{he2020momentum}, MoCo-v2~\cite{chen2020improved}, SimCLR~\cite{chen2020simple}, SimSiam~\cite{chen2021exploring}, and BYOL~\cite{grill2020bootstrap}, as shown in Table \ref{tab:linear classification}. 
We achieved $74.8\%$ and $73.7\%$ top-1 accuracy using R50 and ViT-S, respectively. 
Additionally, OPERA pretrained with $150$ epochs surpasses the performance of the MoCo-v3 baseline as well.
This demonstrates the discriminative ability of the learned representations using OPERA.

\begin{table}[t] \tablesize
    \centering
    \caption{Top-1 accuracy (\%) of the transfer learning on other classification datasets. }
    \vspace{-3mm}
    \setlength\tabcolsep{1.5pt}
    \begin{tabular}{lccccccc}
    \toprule
    Method & P.T. & F.T. & Backbone & C-10 & C-100 & Flowers-102 & Pets \\
    \midrule
    Supervised$\dag$ & 300 & 100 & R50 & 97.6 & 85.5 & 95.6 & 92.2 \\
    MoCo-v3$\dag$ & 300 & 100 & R50 & 97.8 & 86.0 & 93.7 & 90.0 \\
    OPERA & 150 & 100 & R50 & \textbf{97.9} & \textbf{86.3} & \textbf{93.9} & \textbf{91.1} \\
    OPERA & 300 & 100 & R50 & \color{red}\textbf{98.2} & \color{red}\textbf{86.8} & \color{red}\textbf{95.6} & \color{red}\textbf{92.7} \\
    \midrule
    Supervised$\dag$ & 300 & 100 & ViT-S & 98.4 & 86.9 & 95.4 & 93.0 \\
    MoCo-v3$\dag$ & 300 & 100 & ViT-S & 97.9 & 86.6 & 90.3 & 90.1 \\
    OPERA & 150 & 100 & ViT-S & \textbf{98.4} & \textbf{88.5} & \textbf{94.6} & \textbf{91.9} \\
    OPERA & 300 & 100 & ViT-S & \color{red}\textbf{98.6} & \color{red}\textbf{89.0} & \color{red}\textbf{95.5} & \color{red}\textbf{93.3} \\
    \bottomrule
    \end{tabular}
    \vspace{-7mm}
    \label{tab:transfer_cls}
\end{table}

\textbf{End-to-end Finetuning on Imagenet.}
Having pretrained, we finetuned the backbone on the training set of ImageNet. 
We used AdamW~\cite{loshchilov2018decoupled} with an initial learning rate of 5e{-4} and a weight decay of 0.05 and employed the cosine annealing~\cite{loshchilov2016sgdr} learning schedule.
We provide the results in Table \ref{tab:Finetuning} with diverse batch sizes, pretraining epochs, and end-to-end finetuning epochs. 
We see that OPERA consistently achieves better performance under the same setting compared with the MoCo-v3 baseline and DINO~\cite{caron2021emerging}.

\textbf{Transfer to Other Classification Tasks.}
We transferred the pretrained network to other classification tasks including CIFAR-10, CIFAR-100, Oxford Flowers-102, and Oxford-IIIT-Pets. 
We fixed the finetuning epochs to 100 following \cite{chen2021empirical} and reported the top-1 accuracy in Table \ref{tab:transfer_cls}. 
We observe that OPERA obtains better results on four datasets with both R50 and ViT-S. 
Though MoCo-v3 does not show consistent improvement compared to supervised training, our OPERA demonstrates clear superiority.
The results show that OPERA learns generic representations which can widely transfer to smaller classification datasets.

\textbf{Transfer to Semantic Segmentation.}
We also transferred the OPERA-pretrained network to semantic segmentation on ADE20K, which aims at classifying each pixel of an image. 
We adopted MMSegmentaion~\cite{mmseg2020} to conduct the experiments under the same setting.
Specifically, we equipped R50 with FCN~\cite{shelhamer2017fully} and ViTs with UPerNet~\cite{xiao2018unified}. 
We applied SGD~\cite{1985A} with a learning rate of 0.01, a momentum of 0.9, and a weight decay of 5e-4. 
We used a learning schedule of 160k and provided the experimental results in Table \ref{tab:segmentation}. 
We observe consistent improvements over both supervised learning and MoCo-v3 with both R50 and ViTs. 
Particularly, MoCo-v3 performs worse than the supervised model with ViT-S (-0.6 mIoU) while OPERA still outperforms supervised learning with a large margin (+0.9 mIoU).

\begin{table}[t] \tablesize
    \centering
    \caption{Experimental results of semantic segmentation on ADE20K (160k schedule). }
    \vspace{-3mm}
    \setlength\tabcolsep{5pt}
    \begin{tabular}{lccccccc}
    \toprule
    Method & P.T. & Backbone & BS & mIoU & mAcc & aAcc \\
    \midrule
    Supervised & 300 & R50 & 1024 & 36.1 & 45.4 & 77.5\\ 
    MoCo-v3$\dag$ & 300 & R50 & 1024 & 37.0 & 47.0 & 77.6\\
    OPERA & 150 & R50 & 1024 & \textbf{37.7} & \textbf{47.9} & \textbf{77.7}\\
    OPERA & 300 & R50 & 1024 & \textbf{37.9} & \textbf{48.1} & \textbf{77.9}\\
    OPERA & 150 & R50 & 4096 & \textbf{38.1} & \textbf{47.9} & \textbf{78.0}\\
    OPERA & 300 & R50 & 4096 & \color{red}\textbf{38.4} & \color{red}\textbf{48.5} & \color{red}\textbf{78.1}\\
    \midrule
    Supervised & 300 & ViT-S & 1024 & 42.9 & 53.9 & 80.3\\ 
    MoCo-v3$\dag$ & 300 & ViT-S & 1024 & 42.3 & 53.5 & 80.6\\
    OPERA & 150 & ViT-S & 1024 & \textbf{43.4} & \textbf{54.2} & \textbf{80.8}\\
    OPERA & 300 & ViT-S & 1024 & \textbf{43.6} & \textbf{54.4} & \color{red}\textbf{80.9}\\
    OPERA & 150 & ViT-S & 4096 & \textbf{43.5} & \textbf{54.3} & \textbf{80.8}\\
    OPERA & 300 & ViT-S & 4096 & \color{red}\textbf{43.8} & \color{red}\textbf{54.6} & \color{red}\textbf{80.9}\\
    \midrule
    Supervised & 300 & ViT-B & 1024 & 45.4 & 56.5 & 81.4\\ 
    MoCo-v3$\dag$ & 300 & ViT-B & 1024 & 44.4 & 55.1 & 81.5\\
    OPERA & 150 & ViT-B & 1024 & \textbf{44.8} & \textbf{55.7} & \textbf{81.8}\\
    OPERA & 300 & ViT-B & 1024 & \textbf{45.2} & \textbf{55.9} & \textbf{81.9}\\
    MoCo-v3$\dag$ & 300 & ViT-B & 2048 & 45.2 & 55.5 & 81.9\\
    OPERA & 150 & ViT-B & 2048 & \textbf{45.6} & \textbf{56.4} & \textbf{82.0}\\
    OPERA & 300 & ViT-B & 2048 & \textbf{45.9} & \textbf{56.7} & \textbf{82.0}\\
    MoCo-v3$\dag$ & 300 & ViT-B & 4096 & 46.1 & 56.7 & 82.1\\
    OPERA & 150 & ViT-B & 4096 & \textbf{46.4} & \textbf{56.9} & \color{red}\textbf{82.1}\\
    OPERA & 300 & ViT-B & 4096 & \color{red}\textbf{46.6} & \color{red}\textbf{57.2} & \color{red}\textbf{82.1}\\
    \bottomrule
    \end{tabular}
    \vspace{-6mm}
    \label{tab:segmentation}
\end{table}

\textbf{Transfer to Object Detection and Instance Segmentation.}
We further evaluated the transferability of OPERA to object detection and instance segmentation on COCO. 
We performed finetuning and evaluation on COCO$_{train2017}$ and COCO$_{val2017}$, respectively, using the MMDetection~\cite{mmdetection} codebase.
We adopted Mask R-CNN~\cite{he2017mask} with R50-FPN as the detection model. 
We used SGD~\cite{1985A} with a learning rate of 0.02, a momentum of 0.9, and a weight decay of 1e-4.
We reported the performance using the 1 $\times$ schedule (12 epochs) and 2 $\times$ schedule (24 epochs) in Table \ref{tab:detection1x} and Table \ref{tab:detection2x}, respectively. 
We observe that both OPERA and MoCo-v3 demonstrate remarkable advantages compared with random initialization and supervised learning on both object detection and instance segmentation. 
OPERA further improves MoCo-v3 by a relatively large margin on both training schedules, indicating the generalization ability on detection and instance segmentation datasets.

\subsection{Ablation Study}
To further understand the proposed OPERA, we conducted various ablation studies to evaluate its effectiveness.
We mainly focus on end-to-end finetuning on ImageNet for representation discriminativeness and semantic segmentation on ADE20K for representation transferability evaluation on ViT-S.
We fixed the number of finetuning epochs to 100 for ImageNet and used a learning schedule of 160k based on UPerNet~\cite{xiao2018unified} on ADE20K.

\begin{figure*}[t]
\centering
\includegraphics[width=0.96\textwidth]{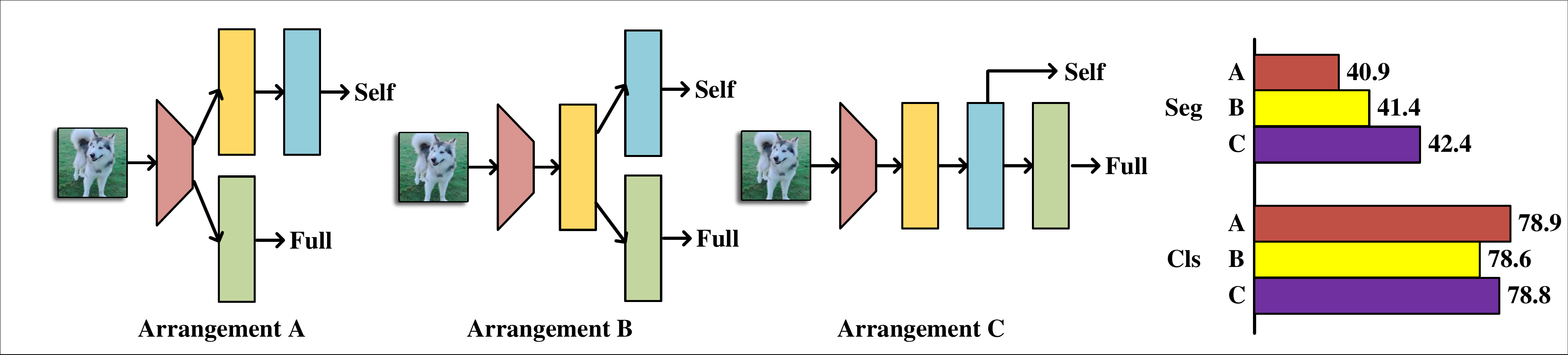}
\vspace{-2mm}
\caption{
Comparisons between different arrangements of supervisions. 
} 
\vspace{-2mm}
\label{fig:arrangement}
\end{figure*}

\textbf{Arrangements of Supervisions.}
As discussed in the paper, the arrangements of supervisions are significant to the quality of the representation.
We thus conducted experiments with different arrangements of supervisions to analyze their effects, as illustrated in \Cref{fig:arrangement}. 
We maintained the basic structure of contrastive learning and impose the fully-supervised training signal on three different positions. 
Note that \Cref{fig:arrangement} only shows the online network of the framework.
Specifically, arrangement A obtains the class-level representation from the backbone and directly imposes the fully-supervised learning signal. 
Differently, arrangement B simultaneously extracts the class-level representation and the instance-level representation with an MLP structure from the projector. 
Arrangement C denotes the proposed OPERA framework in our main experiments.
The experimental results are shown in the right of \Cref{fig:arrangement}.
We observe that arrangement A achieves the highest classification performance on ImageNet. 
This is because the full supervision is directly imposed on the backbone feature, which extracts more class-level information during pretraining. 
However, both arrangements A and B perform much worse on the downstream semantic segmentation task. 
They ignore the underlying hierarchy of the supervisions and do not apply the stronger supervision (full supervision) after the weaker supervision (self-supervision).
The learned representation tends to abandon more instance-level information but obtain more task-specific knowledge, which is not beneficial to the transfer learning tasks. 
Instead, our OPERA (arrangement C) achieves a better balance of class-level and instance-level information learning.

\begin{table}[t] \tablesize
    \centering
    \caption{Experimental results of object detection and instance segmentation on the COCO dataset. (Mask R-CNN, R50-FPN, 1 $\times$ schedule) }
    \vspace{-3mm}
    \renewcommand\arraystretch{1.1}
    \setlength\tabcolsep{1pt}
    \begin{tabular}{lcc|ccc|ccc}
    \hline
    Method & P.T. & BS & $\mathbf{AP}^{bb}$ & $\mathbf{AP}_{50}^{bb}$ & $\mathbf{AP}_{75}^{bb}$ & $\mathbf{AP}^{mk}$ & $\mathbf{AP}_{50}^{mk}$ & $\mathbf{AP}_{75}^{mk}$ \\
    \hline
    Rand. Init. & - & 1024 & 31.0 & 49.5 & 33.2 & 28.5 & 46.8 & 30.4 \\ 
    Supervised & 300 & 1024 & 38.2 & 58.8 & 41.4 & 34.7 & 55.7 & 37.2 \\ 
    \hline
    MoCo-v3$\dag$ & 300 & 1024 & 38.9 & 58.8 & 42.4 & 35.2 & 56.0 & 37.7 \\
    OPERA & 150 & 1024 & \textbf{38.9} & \textbf{58.9} & \textbf{42.1} & \textbf{35.3} & \textbf{55.8} & \textbf{37.8} \\
    OPERA & 300 & 1024 & \textbf{39.2} & \textbf{59.2} & \textbf{42.6} & \textbf{35.9} & \textbf{56.2} & \textbf{38.1} \\
    OPERA & 150 & 4096 & \textbf{39.1} & \textbf{59.1} & \textbf{42.7} & \textbf{35.6} & \textbf{56.2} & \textbf{38.0} \\
    OPERA & 300 & 4096 & \color{red}\textbf{39.3} & \color{red}\textbf{59.3} & \color{red}\textbf{42.9} & \color{red}\textbf{36.0} & \color{red}\textbf{56.4} & \color{red}\textbf{38.1} \\
    \hline
    \end{tabular}
    \vspace{-2mm}
    \label{tab:detection1x}
\end{table}

\begin{table}[t] \tablesize
    \centering
    \caption{Experimental results of object detection and instance segmentation on the COCO dataset (Mask R-CNN, R50-FPN, 2 $\times$ schedule). }
    \vspace{-3mm}
    \renewcommand\arraystretch{1.1}
    \setlength\tabcolsep{1pt}
    \begin{tabular}{lcc|ccc|ccc}
    \hline
    Method & P.T. & BS & $\mathbf{AP}^{bb}$ & $\mathbf{AP}_{50}^{bb}$ & $\mathbf{AP}_{75}^{bb}$ & $\mathbf{AP}^{mk}$ & $\mathbf{AP}_{50}^{mk}$ & $\mathbf{AP}_{75}^{mk}$ \\
    \hline
    Rand. Init. & - & 1024 & 36.7 & 56.7 & 40.0 & 33.7 & 53.8 & 35.9 \\ 
    Supervised & 300 & 1024 & 39.2 & 59.6 & 42.8 & 35.4 & 56.4 & 37.9 \\ 
    \hline
    MoCo-v3$\dag$ & 300 & 1024 & 40.3 & 60.0 & 44.3 & 36.5 & 57.4 & 39.0 \\
    OPERA & 150 & 1024 & \textbf{40.5} & \textbf{60.0} & \textbf{44.6} & \textbf{36.4} & \textbf{57.3} & \textbf{39.0} \\
    OPERA & 300 & 1024 & \textbf{41.2} & \textbf{60.7} & \textbf{45.0} & \textbf{36.9} & \textbf{57.7} & \textbf{39.5} \\
    OPERA & 150 & 4096 & \textbf{41.2} & \textbf{60.9} & \textbf{45.1} & \textbf{37.0} & \textbf{58.0} & \textbf{39.6} \\
    OPERA & 300 & 4096 & \color{red}\textbf{41.5} & \color{red}\textbf{61.2} & \color{red}\textbf{45.5} & \color{red}\textbf{37.3} & \color{red}\textbf{58.2} & \color{red}\textbf{39.9} \\
    \hline
    \end{tabular}
    \vspace{-6mm}
    \label{tab:detection2x}
\end{table}

\newcommand\figwidth{0.22}

\begin{figure*}[t]
\begin{minipage}[t]{\figwidth\textwidth}
  \centering
\includegraphics[width=1\textwidth]{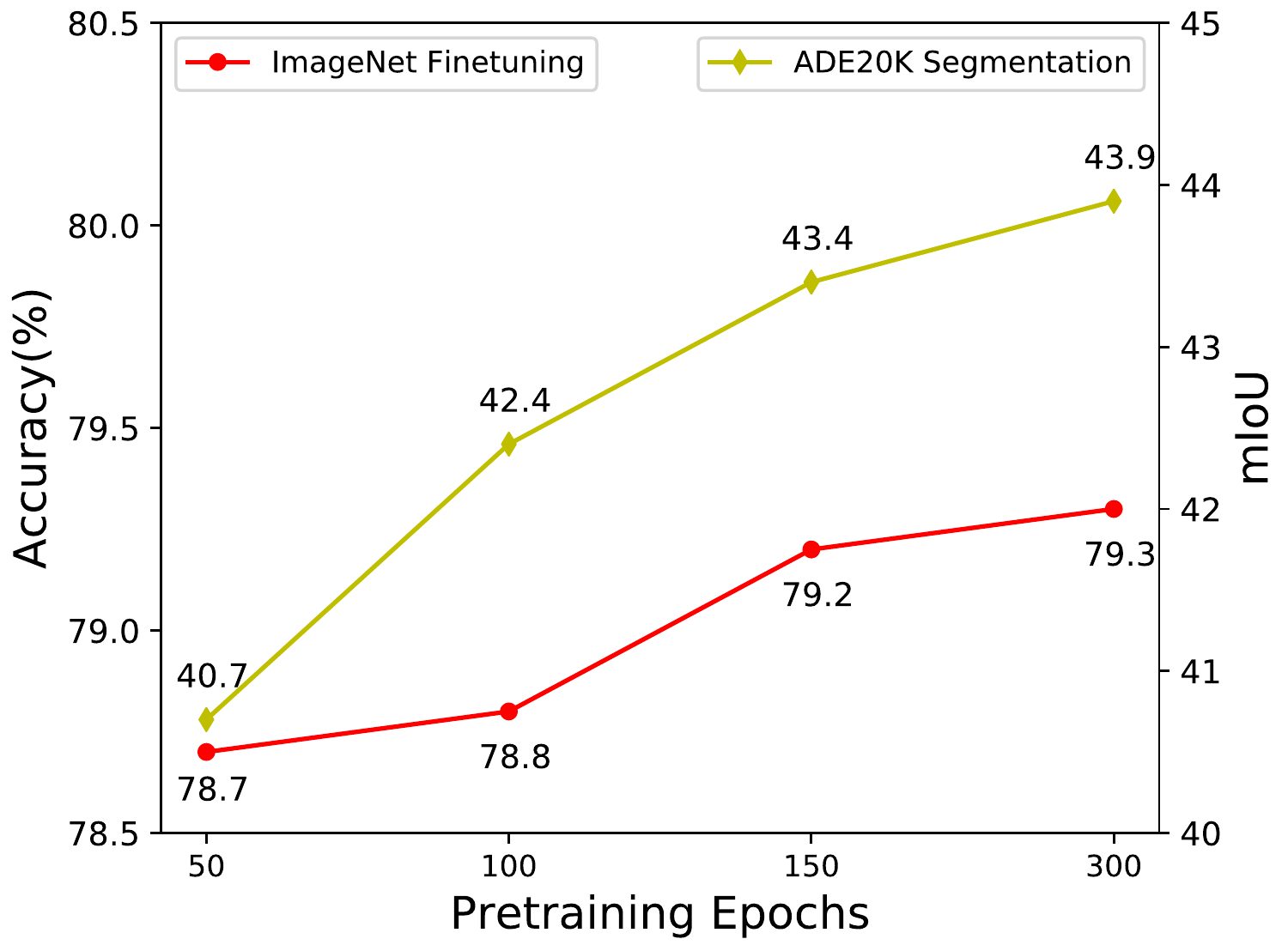}
\vspace{-7mm}
  \caption{
  Effect of 
  pretraining epochs.
  }
\label{fig:pretraining_epoch}
  \end{minipage}
~~~~
\begin{minipage}[t]{\figwidth\textwidth}
  \centering
\includegraphics[width=1\textwidth]{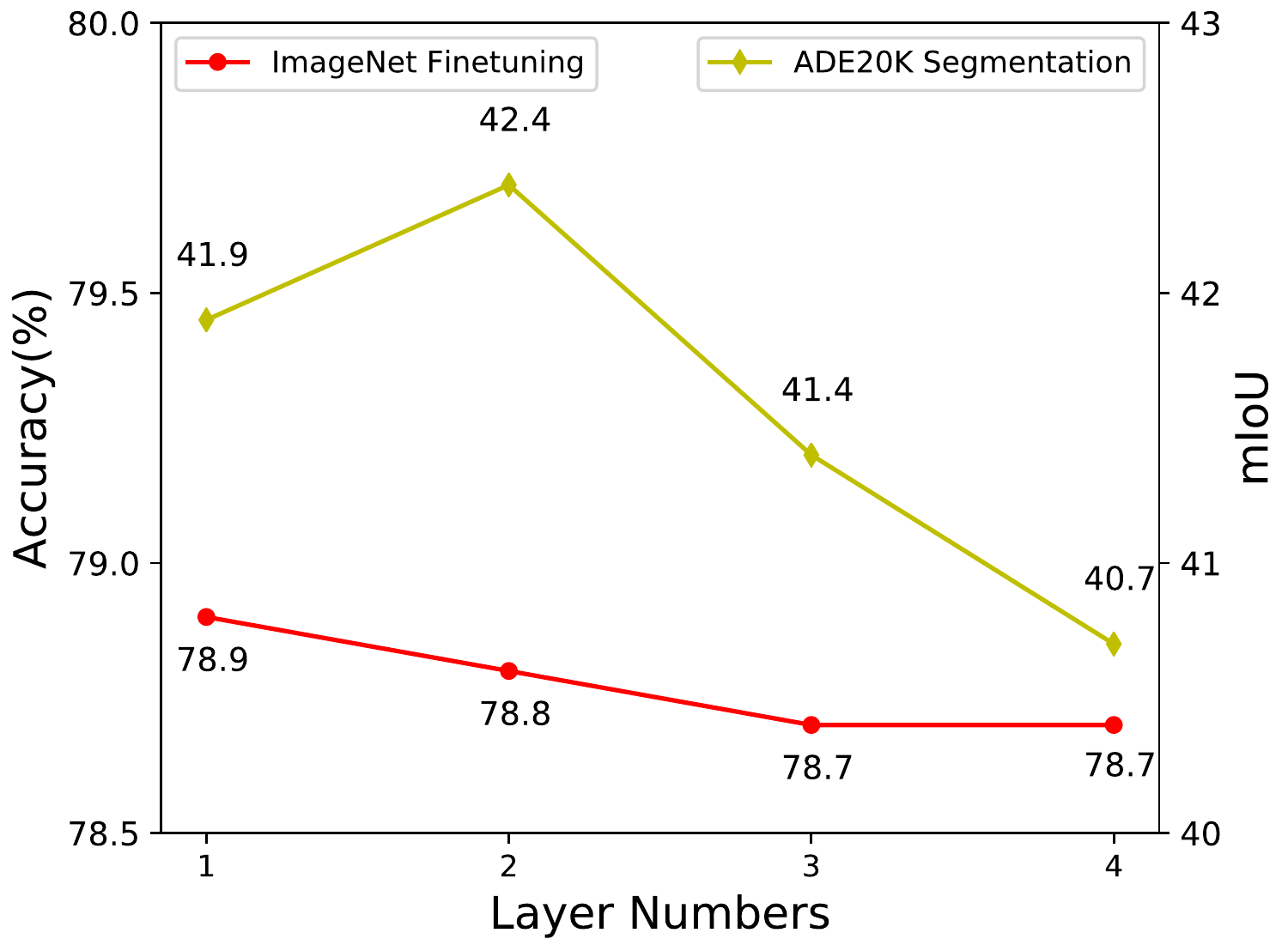}
\vspace{-7mm}
  \caption{
Effect of 
layer numbers of MLP.
}
\label{fig:layer_number}
  \end{minipage}
~~~~
\begin{minipage}[t]{\figwidth\textwidth}
  \centering
\includegraphics[width=1\textwidth]{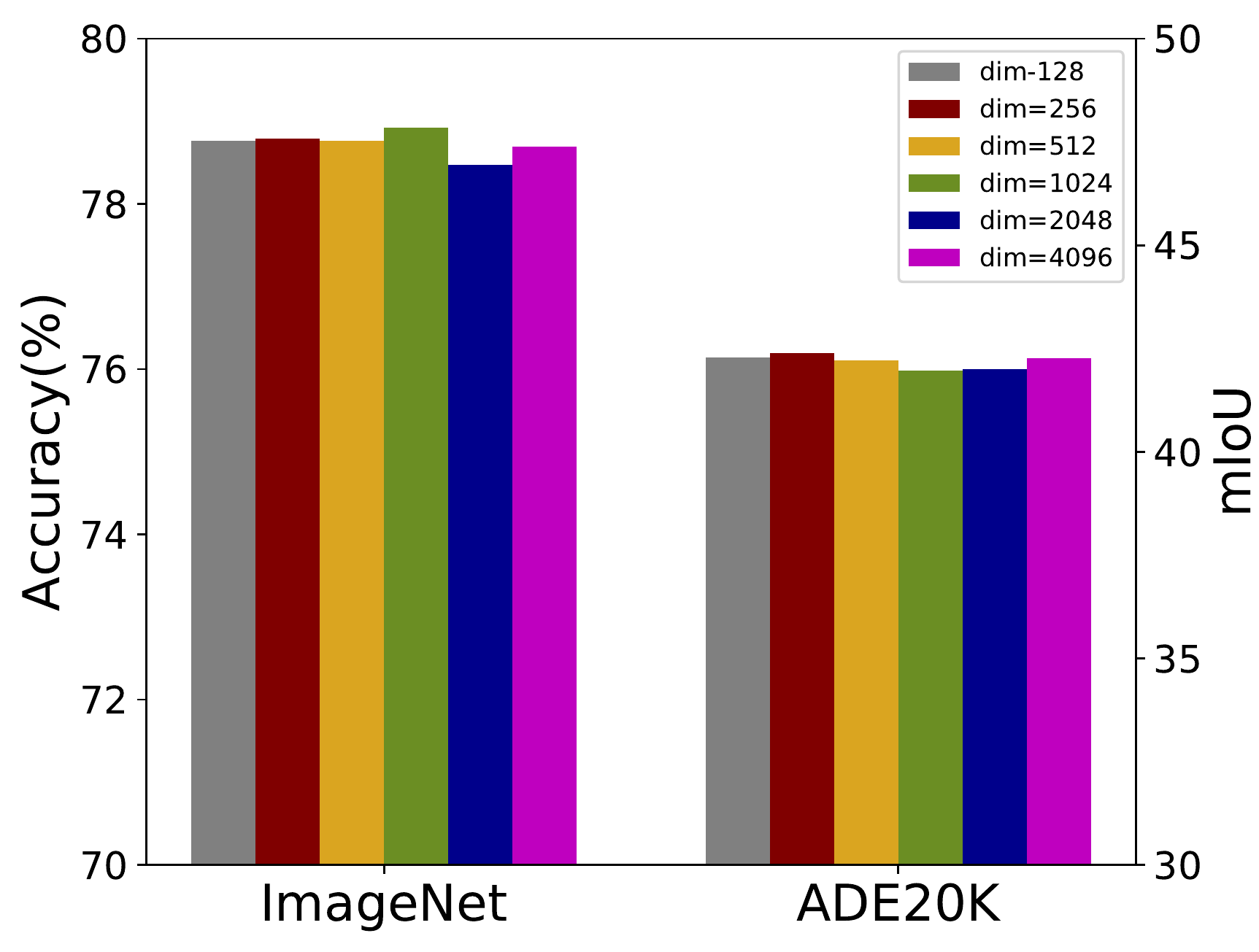}
\vspace{-7mm}
  \caption{
Impact of 
  embedding dimensions.
  }
\label{fig:embed_dim}
  \end{minipage}
~~~~
\begin{minipage}[t]{\figwidth\textwidth}
  \centering
\includegraphics[width=1\textwidth]{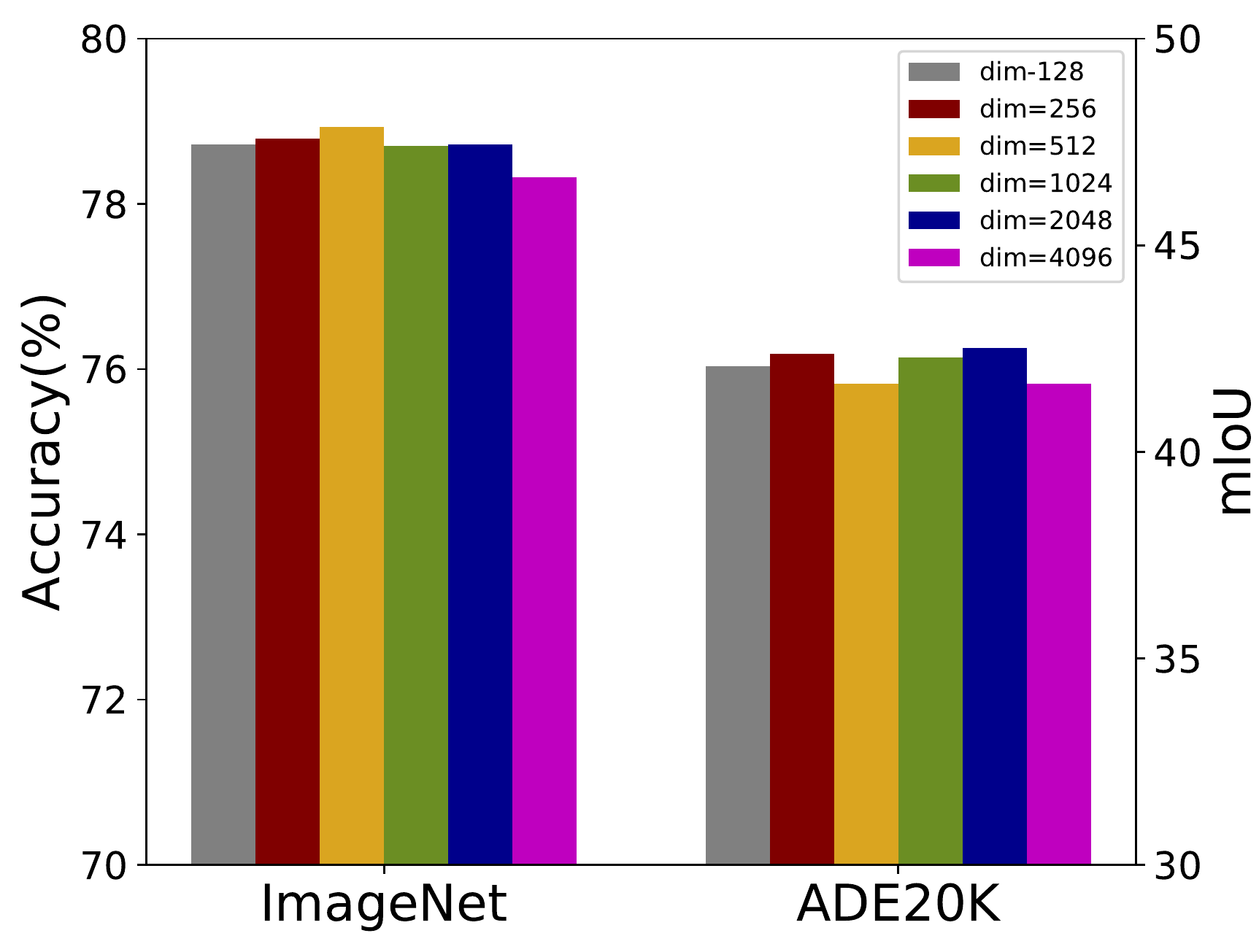}
\vspace{-7mm}
  \caption{
Impact of 
hidden dimensions of MLP.
}
\label{fig:mlp_dim}
  \end{minipage}
\vspace{-7mm}
\end{figure*}

\textbf{Pretraining Epochs.}
We conducted experiments with different pretraining epochs on ImageNet and provided corresponding results in \Cref{fig:pretraining_epoch}. 
We observe that both tasks perform better with longer pretraining epochs. 
Particularly, the performance on semantic segmentation is more sensitive to the number of pretraining epochs compared with ImageNet finetuning, indicating that it takes longer for learning instance-level knowledge.
Note that the finetuning accuracy reaches $78.7\%$ with only 50 pretraining epochs, which demonstrates the efficiency of OPERA.

\textbf{Layer Numbers of MLP.}
We evaluated OPERA with different numbers of fully-connected layers in the final MLP block, as illustrated in \Cref{fig:layer_number}. 
We observe that the classification performance generally decreases with more layers deployed. 
This demonstrates that the class-level supervision is weakened after the MLP block so that the model extracts less class-level information with more layers. 
For semantic segmentation, the mIoU improves (+0.5) when the layer number increases from 1 to 2, indicating that weaker class-level supervision boosts the transferability of the representation.
Still, the performance drops with more layers due to the less effect of the class-level supervision.

\begin{table}[t] \tablesize
    \centering
    \caption{Comparison between supervised pretraining with an MLP projector and OPERA. }
    \vspace{-3mm}
    \setlength\tabcolsep{4pt}
    \begin{tabular}{lcccc}
    \toprule
    Method & P.T. & Backbone. & Top-1 Acc & mIoU \\
    \midrule
    Supervised & 100 & ViT-S & 78.7 & 41.5 \\
    Supervised (MLP) & 100 & ViT-S & 78.4 & 41.9\\
    OPERA & 100 & ViT-S & \textbf{78.8} & \textbf{42.4}\\
    \bottomrule
    \end{tabular}
    \vspace{-5mm}
    \label{tab:supervised pretraining}
\end{table}

\textbf{Embedding Dimensions.}
The embedding dimension in our framework measures the output size of the online network projector. 
We tested the performance using a dimension of 128, 256, 512, 1024, 2048, and 4096 for the embedding and provide the results in \Cref{fig:embed_dim}. 
We observe that the ImageNet accuracy gradually increases before the embedding dimension reaches 512. 
In addition, the model achieves the best segmentation performance when the dimension is 256. 
This indicates that larger dimensions do not necessarily enhance the results because of the information redundancy. 
Therefore, we adopted the embedding dimension of 256 in the main experiments for the best trade-off between model performances and training efficiency.

\textbf{Hidden Dimensions of MLP.}
The hidden dimension of MLP corresponds to the output size of the first linear layer. 
We fixed the other settings and used a dimension of 128, 256, 512, 1024, 2048, and 4096 for comparison, as shown in \Cref{fig:mlp_dim}. 
We see that enlarging the hidden dimension would not necessarily benefit two tasks, indicating that OPERA is not sensitive to the hidden dimensions of MLP. 
Therefore, we employ a dimension of 256 for the main experiments.

\textbf{Transferability for Supervised Learning.}
As illustrated in the previous study~\cite{wang2022revisiting}, adding an MLP block before the classifier of the supervised backbone boosts the transferability of supervised pretraining. 
Therefore, we conducted experiments to compare the performance between the supervised pretraining with an MLP projector and our OPERA framework, as shown in Table~\ref{tab:supervised pretraining}.
We observe that adding the MLP block enhances the transferability for supervised learning while reducing the discriminativenes of the representation.
Nevertheless, OPERA constantly surpasses the discriminativeness and transferability compared with the supervised pretraining with the MLP block, which demonstrates the superiority of the proposed framework.

%% file: chapters/5_conclusion.tex
\section{Conclusion} \label{conclusion}
In this paper, we have presented an omni-supervised representation learning with hierarchical supervisions (OPERA) framework to effectively combine fully-supervised and self-supervised contrastive learning.
We provide a unified perspective of both supervisions and impose the corresponding supervisions on the hierarchical proxy representations in an end-to-end manner. 
We have conducted extensive experiments on classification and other downstream tasks including semantic segmentation and object detection to evaluate the effectiveness of our framework. 
The experimental results have demonstrated the superior classification and transferability of OPERA over both fully supervised learning and self-supervised contrastive learning.
In the future, we will seek to integrate other self-supervised signals such as masked image modeling to further improve the performance.

%% file: chapters/appendix.tex
\section{Proofs}
\subsection{Proof of the Unified Framework} \label{appendix:proof of softmax and infonce}
\begin{proof}
The Softmax loss is formulated as follows:
\begin{equation}\label{equ:cec}
J_s(\mathcal{Y}, \mathcal{P}, \mathcal{L}) = \sum_{\mathbf{y} \in \mathcal{Y}, \mathbf{p} \in \mathcal{P}, l \in \mathcal{L}} -log\frac{exp(s(\mathbf{y},\mathbf{p}))}{\sum_{l_{\mathbf{p'}} \neq l_{\mathbf{y}}} exp(s(\mathbf{y},\mathbf{p'}))},
\end{equation}
where $s(\mathbf{y},\mathbf{p})=\mathbf{y}^T \cdot \mathbf{p}$. 
We compute the gradient of $J_s(\mathcal{Y}, \mathcal{P}, \mathcal{L})$ towards $s(\mathbf{y},\mathbf{p})$ as follows (we omit the summation term for simplicity):
\begin{equation}
\frac{\partial J_s(\mathcal{Y}, \mathcal{P}, \mathcal{L})}{\partial s(\mathbf{y},\mathbf{p})}=
\begin{cases}
-1, &l_{\mathbf{p}}=l_{\mathbf{y}} \\
\frac{exp(s(\mathbf{y},\mathbf{p}))}{\sum_{l_{\mathbf{p'}} \neq l_{\mathbf{y}}} exp(s(\mathbf{y},\mathbf{p'}))}, &l_{\mathbf{p}} \neq l_{\mathbf{y}}
\end{cases}
\end{equation}
Next, we provide the gradient of $J(\mathcal{Y}, \mathcal{P}, \mathcal{L})$ towards $s(\mathbf{y},\mathbf{p})$ for \cref{equ:similarity learning}:
\begin{equation}
\frac{\partial J_s(\mathcal{Y}, \mathcal{P}, \mathcal{L})}{\partial s(\mathbf{y},\mathbf{p})}=
\begin{cases}
-w_p, &l_{\mathbf{p}}=l_{\mathbf{y}} \\
w_n, &l_{\mathbf{p}} \neq l_{\mathbf{y}}
\end{cases}
\end{equation}
Therefore, when we set:
\begin{equation}
	w_p=1, w_n=\frac{exp(s(\mathbf{y},\mathbf{p}))}{\sum_{l_{\mathbf{p'}} \neq l_{\mathbf{y}}} exp(s(\mathbf{y},\mathbf{p'}))},
\end{equation}
the optimization of \cref{equ:similarity learning} is the same as the Softmax loss. 
Thus, we can obtain the softmax objective widely employed in fully supervised learning.

Similarly, when we set \cref{equ:infonce}, the optimization direction of \cref{equ:similarity learning} equals to the InfoNCE loss.
\end{proof}
We refer to Wang~et. al.~\cite{wang2019multi} for more details.

\subsection{Proof of Proposition 1} \label{appendix:proof of prop}
\begin{proof}
    Without loss of generality, we consider the overall supervision on a pair of samples $(\mathbf{y},\mathbf{p})$ in~\cref{equ:hierarchical} as follows:
    \begin{small}
    \begin{equation}\label{equ:supervision}
    \begin{split}
    J^{O}(\mathbf{y},\mathbf{p})&=-I(l_{\mathbf{y}}^{self},l_{\mathbf{p}}^{self}) \cdot w_p^{self} \cdot s(\mathbf{y}^{self},\mathbf{p}^{self}) \\
    &+ (1-I(l_{\mathbf{y}}^{self},l_{\mathbf{p}}^{self})) \cdot w_n^{self} \cdot s(\mathbf{y}^{self},\mathbf{p}^{self})\\
    & -I(l_{\mathbf{y}}^{full},l_{\mathbf{p}}^{full}) \cdot w_p^{full} \cdot s(\mathbf{y}^{full},\mathbf{p}^{full}) \\
    &+ (1-I(l_{\mathbf{y}}^{full},l_{\mathbf{p}}^{full})) \cdot w_n^{full} \cdot s(\mathbf{y}^{full},\mathbf{p}^{full})
    \end{split}
    \end{equation}
    \end{small}
    We then compute the gradient of $J^{O}(\mathbf{y},\mathbf{p})$ towards $\mathbf{y}$ as follows:
    \begin{small}
    \begin{equation}\label{equ:gradient}
    \begin{split}
    \frac{\partial J^{O}(\mathbf{y},\mathbf{p})}{\partial \mathbf{y}} &= -I(l_{\mathbf{y}}^{self},l_{\mathbf{p}}^{self}) \cdot w_p^{self} \cdot \mW_g^T \gamma(\mathbf{y}^{self}, \mathbf{p}^{self}) \\
    &+ (1-I(l_{\mathbf{y}}^{self},l_{\mathbf{p}}^{self})) \cdot w_n^{self} \cdot \mW_g^T \gamma(\mathbf{y}^{self}, \mathbf{p}^{self})\\
    &-I(l_{\mathbf{y}}^{full},l_{\mathbf{p}}^{full}) \cdot w_p^{full} \cdot \mW_g^T\mW_h^T \gamma(\mathbf{y}^{full}, \mathbf{p}^{full}) \\
    & + (1-I(l_{\mathbf{y}}^{full},l_{\mathbf{p}}^{full})) \cdot w_n^{full} \cdot \mW_g^T\mW_h^T \gamma(\mathbf{y}^{full}, \mathbf{p}^{full})
    \end{split}
    \end{equation}
    \end{small}
where $\gamma(\mathbf{y}, \mathbf{p}_p)=\frac{\partial s(\mathbf{y},\mathbf{p}_p)}{\partial \mathbf{y}}$. For simplicity and clarity, we define $s(\mathbf{y},\mathbf{p})=\mathbf{y}^T\mathbf{p}$. Under such circumstances,~\cref{equ:gradient} can be formulated as follows:
    \begin{small}
     \begin{equation}\label{equ:gradient1}
    \begin{split}
    \frac{\partial J^{O}(\mathbf{y},\mathbf{p})}{\partial \mathbf{y}} &= -I(l_{\mathbf{y}}^{self},l_{\mathbf{p}}^{self}) \cdot w_p^{self} \cdot \mW_g^T \mW_g \mathbf{p} \\
    &+ (1-I(l_{\mathbf{y}}^{self},l_{\mathbf{p}}^{self})) \cdot w_n^{self} \cdot \mW_g^T \mW_g \mathbf{p}\\
    &-I(l_{\mathbf{y}}^{full},l_{\mathbf{p}}^{full}) \cdot w_p^{full} \cdot \mW_g^T\mW_h^T \mW_h \mW_g \mathbf{p} \\
    &+ (1-I(l_{\mathbf{y}}^{full},l_{\mathbf{p}}^{full})) \cdot w_n^{full} \cdot \mW_g^T\mW_h^T \mW_h \mW_g \mathbf{p}
    \end{split}
    \end{equation}
    \end{small}
The concrete form of~\cref{equ:gradient1} is determined by the label connection between $\mathbf{y}$ and $\mathbf{p}$. Specifically, when $I(l_{\mathbf{y}}^{self},l_{\mathbf{p}}^{self})\cdot I(l_{\mathbf{y}}^{full},l_{\mathbf{p}}^{full})=1$, denoting that $\mathbf{y}$ and $\mathbf{p}$ shares the same self-supervised and fully supervised label,~\cref{equ:gradient1} degenerates to:
\begin{equation}
\begin{split}
\frac{\partial J^{O}(\mathbf{y},\mathbf{p})}{\partial \mathbf{y}} = \mW_g^T(-w_p^{self}\mI-w_p^{full}\mW_h^T\mW_h)\mW_g \mathbf{p} 
\end{split}    
\end{equation}
Similarly, when $(1-I(l_{\mathbf{y}}^{self},l_{\mathbf{p}}^{self}))\cdot I(l_{\mathbf{y}}^{full},l_{\mathbf{p}}^{full})=1$,~\cref{equ:gradient1} degenerates to:
\begin{equation}\label{equ:conflictexample}
\begin{split}
\frac{\partial J^{O}(\mathbf{y},\mathbf{p})}{\partial \mathbf{y}} = \mW_g^T(w_n^{self}\mI-w_p^{full}\mW_h^T\mW_h)\mW_g \mathbf{p}
\end{split}    
\end{equation}
And when $(1-I(l_{\mathbf{y}}^{self},l_{\mathbf{p}}^{self}))\cdot (1-I(l_{\mathbf{y}}^{full},l_{\mathbf{p}}^{full}))=1$,~\cref{equ:gradient1} degenerates to:
\begin{equation}
\begin{split}
\frac{\partial J^{O}(\mathbf{y},\mathbf{p})}{\partial \mathbf{y}} = \mW_g^T(w_n^{self}\mI+w_n^{full}\mW_h^T\mW_h)\mW_g \mathbf{p}
\end{split}    
\end{equation}
Next, we consider that $\mathbf{p}$ is fixed during optimization (e.g., a prototype) and provide the differential of $s(\mathbf{y}, \mathbf{p})$ based on~\cref{equ:conflictexample}:
\begin{equation}
\begin{split}
\Delta s^{O}(\mathbf{y},\mathbf{p}) &\propto (\frac{\partial J^{O}(\mathbf{y},\mathbf{p})}{\partial \mathbf{y}})^T \cdot \mathbf{p} \\
&= \mathbf{p}^T \mW_g^T(w_n^{self}\mI-w_p^{full}\mW_h^T\mW_h)\mW_g \mathbf{p} \\
&=w_n^{self}(\mathbf{p}^{self})^T \mathbf{p}^{self} - w_p^{full} (\mathbf{p}^{full})^T \mathbf{p}^{full} \\
&= w_n^{self}\alpha(\mW_g)-w_p^{full}\beta(\mW_g,\mW_h),
\end{split}
\end{equation}
where $\alpha(\mW_g) = (\mathbf{p}^{self})^T \mathbf{p}^{self}$ and $\beta(\mW_g,\mW_h) = (\mathbf{p}^{full})^T \mathbf{p}^{full}$.
Therefore, we formulate the above equation considering all the possible relations between the label of $\mathbf{y}$ and $\mathbf{p}$ as follows:
\begin{equation}
\begin{split}
\Delta s^{O}(\mathbf{y},\mathbf{p})&\propto I(l_{\mathbf{y}}^{self},l_{\mathbf{p}}^{self})\cdot I(l_{\mathbf{y}}^{full},l_{\mathbf{p}}^{full}) \\
&\cdot (-w_p^{self}\alpha(\mW_g)-w_p^{full}\beta(\mW_g,\mW_h))\\ 
+ &(1-I(l_{\mathbf{y}}^{self},l_{\mathbf{p}}^{self}))\cdot I(l_{\mathbf{y}}^{full},l_{\mathbf{p}}^{full})\\
&\cdot (w_n^{self}\alpha(\mW_g)-w_p^{full}\beta(\mW_g,\mW_h)) \\
+ &(1-I(l_{\mathbf{y}}^{self},l_{\mathbf{p}}^{self}))\cdot (1-I(l_{\mathbf{y}}^{full},l_{\mathbf{p}}^{full})) \\
&\cdot(w_n^{self}\beta(\mW_g) + w_n^{full}\alpha(\mW_g,\mW_h))
\end{split}    
\end{equation}
For~\cref{equ:equivalent}, we similarly consider a pair of samples $(\mathbf{y},\mathbf{p})$ and we can obtain the gradient of $J(\mathbf{y},\mathbf{p})$ towards $s(\mathbf{y},\mathbf{p})$ as follows:
\begin{equation}\label{equ:equivalent_gradient}
\begin{split}
\frac{\partial J(\mathbf{y},\mathbf{p})}{\partial s(\mathbf{y},\mathbf{p})} &=I(l_{\mathbf{y}}^{self},l_{\mathbf{p}}^{self})\cdot I(l_{\mathbf{y}}^{full},l_{\mathbf{p}}^{full})\\
& \cdot (-w_p^{self}\alpha(\mW_g)-w_p^{full}\beta(\mW_g,\mW_h))\\ 
&+ (1-I(l_{\mathbf{y}}^{self},l_{\mathbf{p}}^{self}))\cdot I(l_{\mathbf{y}}^{full},l_{\mathbf{p}}^{full})\\
&\cdot (w_n^{self}\alpha(\mW_g)-w_p^{full}\beta(\mW_g,\mW_h)) \\
&+ (1-I(l_{\mathbf{y}}^{self},l_{\mathbf{p}}^{self}))\cdot (1-I(l_{\mathbf{y}}^{full},l_{\mathbf{p}}^{full}))\\
&\cdot(w_n^{self}\beta(\mW_g) + w_n^{full}\alpha(\mW_g,\mW_h))
\end{split}    
\end{equation}
The differential of $s(\mathbf{y},\mathbf{p})$ during optimization for~\cref{equ:equivalent} is proportional to to $\frac{\partial J(\mathbf{y},\mathbf{p})}{\partial s(\mathbf{y},\mathbf{p})}$:
\begin{equation}
\begin{split}
\Delta s(\mathbf{y},\mathbf{p})&\propto I(l_{\mathbf{y}}^{self},l_{\mathbf{p}}^{self})\cdot I(l_{\mathbf{y}}^{full},l_{\mathbf{p}}^{full}) \\
&\cdot (-w_p^{self}\alpha(\mW_g)-w_p^{full}\beta(\mW_g,\mW_h))\\ 
&+ (1-I(l_{\mathbf{y}}^{self},l_{\mathbf{p}}^{self}))\cdot I(l_{\mathbf{y}}^{full},l_{\mathbf{p}}^{full})\\
&\cdot (w_n^{self}\alpha(\mW_g)-w_p^{full}\beta(\mW_g,\mW_h)) \\
&+ (1-I(l_{\mathbf{y}}^{self},l_{\mathbf{p}}^{self}))\cdot (1-I(l_{\mathbf{y}}^{full},l_{\mathbf{p}}^{full}))\\
&\cdot(w_n^{self}\beta(\mW_g) + w_n^{full}\alpha(\mW_g,\mW_h))
\end{split}    
\end{equation}
Therefore, the optimization towards $s(\mathbf{y},\mathbf{p})$ of~\cref{equ:equivalent} is equal to~\cref{equ:hierarchical}.
In addition, this conclusion is also applicable to the summation form of~\cref{equ:equivalent} and~\cref{equ:hierarchical}, which means that~\cref{equ:equivalent} is an equivalent form of~\cref{equ:hierarchical}.
\end{proof}

\subsection{Proof of Corollary 1} \label{appendix:proof of coro1}
\begin{proof}
With the gradient of~\cref{equ:equivalent} in~\cref{equ:equivalent_gradient}, we provide the loss weight on $(\mathbf{y},\mathbf{p})$ as follows:
\begin{small}
\begin{equation}
w(l_{\mathbf{y}}^{self}=l_{\mathbf{p}}^{self},l_{\mathbf{y}}^{full}=l_{\mathbf{p}}^{full}) = -w_p^{self}\alpha(\mW_g)-w_p^{full}\beta(\mW_g,\mW_h)
\end{equation}
\begin{equation} \label{equ:contra}
w(l_{\mathbf{y}}^{self}\neq l_{\mathbf{p}}^{self},l_{\mathbf{y}}^{full}=l_{\mathbf{p}}^{full}) = w_n^{self}\alpha(\mW_g)-w_p^{full}\beta(\mW_g,\mW_h)
\end{equation}
\begin{equation}
w (l_{\mathbf{y}}^{self}\neq l_{\mathbf{p}}^{self}, l_{\mathbf{y}}^{full}\neq l_{\mathbf{p}}^{full}) = w_n^{self}\alpha(\mW_g) + w_n^{full}\beta(\mW_g,\mW_h)
\end{equation}
\end{small}
Therefore, we can obtain the following two inequalities:
\begin{equation}
\begin{split}
&w(l_{\mathbf{y}}^{self}=l_{\mathbf{p}}^{self},l_{\mathbf{y}}^{full} =l_{\mathbf{p}}^{full}) - w(l_{\mathbf{y}}^{self}\neq l_{\mathbf{p}}^{self},l_{\mathbf{y}}^{full}=l_{\mathbf{p}}^{full}) \\
&= -w_p^{self}\alpha(\mW_g) - w_n^{self}\alpha(\mW_g) \leq 0
\end{split}
\end{equation}
\begin{equation}
\begin{split}
& w(l_{\mathbf{y}}^{self}\neq l_{\mathbf{p}}^{self},l_{\mathbf{y}}^{full}=l_{\mathbf{p}}^{full})- w (l_{\mathbf{y}}^{self}\neq l_{\mathbf{p}}^{self}, l_{\mathbf{y}}^{full}\neq l_{\mathbf{p}}^{full}) \\
&= w_p^{full}\beta(\mW_g,\mW_h) - w_n^{full}\beta(\mW_g,\mW_h) \leq 0
\end{split}
\end{equation}
We organize the above inequalities and can obtain:
\begin{equation}
\begin{split}
	& w(l_{\mathbf{y}}^{self}=l_{\mathbf{p}}^{self},l_{\mathbf{y}}^{full}=l_{\mathbf{p}}^{full}) \\
	& \leq w(l_{\mathbf{y}}^{self}\neq l_{\mathbf{p}}^{self},l_{\mathbf{y}}^{full}=l_{\mathbf{p}}^{full}) \\
	& \leq w (l_{\mathbf{y}}^{self}\neq l_{\mathbf{p}}^{self}, l_{\mathbf{y}}^{full}\neq l_{\mathbf{p}}^{full}).
\end{split}
\end{equation}

\end{proof}

\subsection{Proof of Corollary 2} \label{appendix:proof of coro2}
\begin{proof}
For the contradictory situation, i.e., $I(l_{\mathbf{y}}^{self}, l_{\mathbf{p}}^{self}) = 0$ and $I(l_{\mathbf{y}}^{full}, l_{\mathbf{p}}^{full}) = 1$, the loss weight is the same as \cref{equ:contra}:
\begin{small}
\begin{equation}
w(l_{\mathbf{y}}^{self}\neq l_{\mathbf{p}}^{self},l_{\mathbf{y}}^{full}=l_{\mathbf{p}}^{full}) = w_n^{self} \cdot  \alpha(\mW_g) - w_p^{full} \cdot  \beta(\mW_g,\mW_h).
\end{equation}
\end{small}
\end{proof}
The direction and intensity of optimization is determined by the values of $\alpha(\mW_g)$ and $\beta(\mW_g,\mW_h)$. For example, when $w_n^{self} \cdot  \alpha(\mW_g) - w_p^{full} \cdot  \beta(\mW_g,\mW_h) < 0$, the model increases the similarity between $\mathbf{y}$ and $\mathbf{p}$ during optimization. Consequently, OPERA adaptively adjusts the loss weight between each pair of samples to resolve the contradiction in~\cref{equ:naive}.

\section{Implementation Details}
We provide more implementation details of our experiments on linear evaluation, end-to-end finetuning, semantic segmentation, and object detection.

\subsection{Linear Evaluation and End-to-End Finetuning}
We evaluated our method on linear evaluation and end-to-end finetuning on the ImageNet~\cite{russakovsky2015imagenet} dataset. 
For linear evaluation, we used the SGD optimizer and fixed the batch size to 1024. 
We set the learning rate to 0.1 for R50~\cite{he2016deep} and 3.0 for DeiT-S~\cite{touvron2021training}.
The weight decay was 0 and the momentum of the optimizer was 0.9 for both architectures. 
Additionally, we conducted end-to-end finetuning with DeiTs and respectively set the batch size to 1024, 2048, and 4096. 
We used the AdamW~\cite{loshchilov2018decoupled} optimizer with an initial learning rate of 5e{-4} and a weight decay of 0.05. 
We employed the cosine annealing~\cite{loshchilov2016sgdr} learning schedule during training.

\subsection{Semantic Segmentation}
We transferred the pretrained models to the semantic segmentation task with R50 and DeiTs on the ADE20K~\cite{zhou2019semantic} dataset. 
For R50, we used FCN~\cite{shelhamer2017fully} as the basic segmentation head. 
We applied the SGD~\cite{1985A} optimizer with a learning rate of 0.01, a momentum of 0.9, and a weight decay of 5e-4. 
For DeiTs, we adopted the UperNet~\cite{xiao2018unified} as the basic decoder and FCN~\cite{shelhamer2017fully} as the auxiliary head. 
The optimizer, the momentum, and the weight decay are the same as R50. In addition, we trained the models for 160k for both architectures.

\subsection{Object Detection}
We conducted experiments on object detection and instance segmentation with R50 on the COCO~\cite{lin2014microsoft} dataset.
We employed Mask R-CNN~\cite{he2017mask} with R50-FPN as the backbone. 
We used the SGD~\cite{1985A} optimizer with a learning rate of 0.02, a momentum of 0.9, and a weight decay of 1e-4 for both 1 $\times$ and 2 $\times$ schedules.

\begin{table}[t] \tablesize
    \centering
    \caption{Top-1 accuracy (\%) under the end-to-end finetuning protocol on ImageNet based on MIM methods. }
    \vspace{-3mm}
    \setlength\tabcolsep{1pt}
    \begin{tabular}{lcccc}
    \toprule
    Method & Type & Pretraining & Backbone & Top-1 Acc \\
    \midrule
    BEiT & Masked Image Modeling & 800 & ViT-B & 83.2\\
    MSN & Masked Image Modeling & 600 & ViT-B & 83.4\\
    MAE & Masked Image Modeling & 1600 & ViT-B & 83.6\\
    iBOT & Masked Image Modeling & 1600 & ViT-B & 83.8\\
    SimMIM & Masked Image Modeling & 800 & ViT-B & 83.8\\
    \midrule
    DINO$\dag$ & Contrastive Learning & 300 & ViT-B & 82.8 \\
    MoCo-v3$\dag$ & Contrastive Learning & 300 & ViT-B & 83.0\\
    OPERA & Contrastive Learning & 300 & ViT-B & 83.5\\
    \bottomrule
    \end{tabular}
    \vspace{-5mm}
    \label{tab:MIM}
\end{table}

\section{Generalizing to MIM Methods}
The recent emergence of a new type of self-supervised learning method, masked image modeling (MIM), has demonstrated promising results on vision transformers. 
MIM masks part of the input images and aims to reconstruct the masked parts of the image.
It extracts the representations based on the masked images and uses reconstruction as the objective to learn meaningful representations.
For example, MAE~\cite{he2021masked} adopts an encoder to extract the representations of unmasked tokens and a decoder to reconstruct the whole image with the representations. 
MIM-based methods typically outperform existing self-supervised contrastive learning methods by a large margin~\cite{he2021masked} on ViTs as shown in~\cref{tab:MIM}.
We show several MIM-based methods including BEiT~\cite{bao2021beit}, MSN~\cite{assran2022masked}, MAE~\cite{he2021masked}, iBOT~\cite{zhou2021ibot}, and SimMIM~\cite{xie2022simmim}. 
We see that MIM-based methods tend to pretrain the models for more epochs and obtain better performances than contrastive learning methods. 
Though OPERA fails to achieve better performance than all MIM-based methods, the gap is further reduced with fewer training epochs required.
Particularly, our OPERA framework achieves $83.5\%$ top-1 accuracy and is comparable with MIM-based methods (even higher than BEiT~\cite{bao2021beit} and MSN~\cite{assran2022masked}), which demonstrates the effectiveness of the proposed method. 

As an interesting future work, OPERA can be easily extended to MIM by inserting a new task space in our hierarchy.
As MIM aims to reconstruct a specific view of an instance, we deem that it learns more low-level features than self-supervised contrastive learning (instance-level).
Therefore, we expect to insert the task space of MIM below the self-supervised contrastive learning space:
\begin{equation}
\mathcal{Y}^{mask}=\mathcal{Y}, \quad \mathcal{Y}^{self}=g(\mathcal{Y}), \quad \mathcal{Y}^{full}=h(\mathcal{Y}^{self}).
\end{equation}
The overall objective of OPERA is then:
\begin{equation}\label{equ:OPERA_MIM}
\begin{split}
J^{O}(\mathcal{Y},\mathcal{P},\mathcal{L})&=J^{mask}(\mathcal{Y}^{mask},\mathcal{L}^{mask})\\
& + J^{self}(\mathcal{Y}^{self},\mathcal{P}^{self},\mathcal{L}^{self}) \\
&+ J^{full}(\mathcal{Y}^{full},\mathcal{P}^{full},\mathcal{L}^{full}),
\end{split}  
\end{equation}
where $J^{mask}(\mathcal{Y}^{mask},\mathcal{L}^{mask})$ is the MIM learning objective.
We leave the experiments with \cref{equ:OPERA_MIM} as future works.